\documentclass[11pt,a4paper]{article}

\usepackage[margin=1in]{geometry}
\usepackage{microtype}
\usepackage{ragged2e}   
\usepackage[table,xcdraw]{xcolor}  

\usepackage{amsmath, amssymb, amsfonts}
\usepackage{booktabs}
\usepackage{multirow}
\usepackage{tabularx}
\usepackage{longtable}
\usepackage{graphicx}
\usepackage{caption}
\usepackage{subcaption}
\usepackage{float}
\usepackage[most]{tcolorbox}

\usepackage{cite}        
\usepackage{url}
\usepackage{hyperref}
\hypersetup{
  colorlinks=true,
  linkcolor=blue,
  citecolor=blue,
  urlcolor=blue
}

\usepackage{times}
\usepackage{soul}
\usepackage{enumitem}
\usepackage{xspace}      
\usepackage{amsthm}

\usepackage[linesnumbered,ruled,vlined]{algorithm2e}

\usepackage{authblk}

\newcommand{\ModelName}{\textsc{\textit{MedBayes-Lite}}\xspace}

\makeatletter
\let\oldthebibliography\thebibliography

\renewcommand{\thebibliography}[1]{%
  \oldthebibliography{#1}%
  \setcounter{enumiv}{0}
}
\makeatother

\title{\vspace{-1cm}\textbf{\Large MedBayes-Lite: A Clinical Uncertainty Governance Layer for Risk-Aware Medical Decision Support}\vspace{0.3cm}}

\author[1,*]{\normalsize Elias Hossain}
\author[2]{\normalsize Md Mehedi Hasan Nipu}
\author[3]{\normalsize Maleeha Sheikh}
\author[1]{\normalsize Tasfia Nuzhat}
\author[4]{\normalsize Rajib Rana}
\author[5]{\normalsize Subash Neupane}
\author[6,7]{\normalsize Björn W. Schuller}
\author[1]{\normalsize Niloofar Yousefi}

\affil[1]{\footnotesize College of Engineering and Computer Science, University of Central Florida, Orlando, FL 32816, USA}
\affil[2]{\footnotesize Department of Computer Science and Engineering, North South University, Dhaka 1229, Bangladesh}
\affil[3]{\footnotesize Department of Electrical and Computer Engineering, Purdue University Fort Wayne, Fort Wayne, IN 46805, USA}
\affil[4]{\footnotesize School of Mathematics, Physics and Computing, University of Southern Queensland, Springfield Central, QLD 4300, Australia}
\affil[5]{\footnotesize Meharry Medical College, Nashville, TN 37208, USA}
\affil[6]{\footnotesize CHI -- Chair of Health Informatics, Technical University of Munich (TUM), Munich, Germany}
\affil[7]{\footnotesize GLAM -- Group on Language, Audio, \& Music, Imperial College London, London, UK}

\affil[*]{\footnotesize Corresponding Author: \texttt{mdelias.hossain@ucf.edu}}

\date{}

\begin{document}
\justifying

\maketitle

\begin{abstract}
Clinical language models can be overconfident, assigning high confidence to incorrect predictions, especially on high-severity and out-of-distribution benchmark cases. We present \textit{MedBayes-Lite}\footnote{Source code and experimental implementation are publicly available at: \url{https://github.com/eliashossain001/medbayes-lite}.}, a clinical uncertainty governance layer that converts inference-time uncertainty into a risk-aware decision policy for transformer-based clinical predictors. The layer combines Monte Carlo dropout uncertainty, calibration of the predictive distribution, and a confidence-guided abstention gate that defers low-confidence predictions for human review. It requires no retraining and adds no trainable parameters, and it applies directly to open-weight encoder models. A limited sampling-only generative contrast using Qwen2.5-7B-Instruct is included solely for qualitative comparison. We evaluate the framework as a governance mechanism rather than as a ranking method. On clinical multiple-choice question answering (MedMCQA and MedQA-USMLE), the layer reduces expected calibration error by 0.23 to 0.33 absolute and drives \emph{harmful} overconfident errors, those that are confident, incorrect, and clinically high-severity, toward zero ($p\approx0$). The benefit is strongest under medical domain shift: when models trained on MedMCQA are evaluated on MedQA-USMLE, an uncalibrated baseline commits confident errors on about 21\% of high-severity items, and the governance layer reduces this to near zero while roughly halving the calibration drift. We introduce and empirically evaluate the Clinical Uncertainty Score (CUS), a severity-weighted calibration metric that correlates strongly with harmful overconfidence ($r\approx0.88$). We also report negative results honestly: the framework does not improve the area under the risk-coverage curve, temperature scaling matches its calibration at lower cost, deep ensembles achieve better risk ranking, the safety gains carry a measurable coverage cost, and on already well-calibrated models the layer provides no benefit. A contrast with an open-weight generative model (Qwen2.5-7B-Instruct) shows that naive self-consistency does not confer safety, which further motivates explicit uncertainty governance. The reported experiments evaluate the calibration and confidence-guided abstention components; the optional uncertainty-weighted attention pathway remained inactive ($\lambda=0$). We position \textit{MedBayes-Lite} not as a new uncertainty-aware architecture, but as a deployable calibration-and-abstention layer that reduces confident high-severity errors on clinical question-answering benchmarks.
\end{abstract}

\section{Introduction}
\label{sec:introduction}

Transformer-based language models are increasingly used in clinical decision support, for symptom triage and preliminary differential diagnosis \cite{ong2025large,nerella2024transformers,arslan2025evaluating,mcduff2025towards}, clinical note drafting \cite{oliveira2025development}, medication reconciliation and dosing checks \cite{sridharan2024unlocking}, retrieval and synthesis of guideline-based recommendations \cite{li2025streamlining}, and patient-facing communication~\cite{lu2024large,alberts2023large,thirunavukarasu2023large,neupane2025medinsight,neupane2024clinicsum}. As these systems move toward routine use, a specific safety failure becomes central: they are frequently \emph{confidently wrong}. Modern clinical predictors are poorly calibrated, and their miscalibration is worst exactly where the cost of error is highest, on ambiguous, high-severity, and out-of-distribution cases \cite{lievin2024can,savage2024large,mora2024trustworthy}. A single confidently incorrect recommendation in triage, diagnosis, or dosing can cascade into harm, and an automated answer engine that never signals doubt invites automation bias \cite{salvi2025explainability,catak2024uncertainty}. The central risk is therefore not only factual error, but the absence of a mechanism that recognizes when a prediction should not be acted upon.

Clinical decision-making is inherently probabilistic. Uncertainty arises from measurement noise and physiological variability, incomplete or missing evidence, symptom-to-disease ambiguity, conflicting findings across modalities or time, and documentation noise. Bayesian modeling organizes these into two classes: \textbf{aleatoric} uncertainty, the irreducible randomness in observations and outcomes, and \textbf{epistemic} uncertainty, the model and knowledge uncertainty due to limited data, domain shift, or mis-specification~\cite{hullermeier2021aleatoric,abdar2021review}. Clinicians respond to uncertainty by slowing down, ordering additional tests, and escalating to expert consultation. Current clinical models rarely expose such signals or act on them, and instead collapse weak, missing, or contradictory evidence into overconfident outputs.

Existing uncertainty quantification offers partial answers but leaves a governance gap. Post-hoc calibration, such as temperature scaling and isotonic regression, rescales reported confidence; it is cheap and effective, but it adjusts probabilities without changing what the model does with them, and it can degrade under distribution shift~\cite{10.5555/3305381.3305518,niculescu2005predicting}. Ensemble and Bayesian methods estimate epistemic uncertainty more directly but are computationally heavy~\cite{lakshminarayanan2017simple,maddox2019simple}. Crucially, most of this work stops at \emph{estimating} uncertainty and reports it as a number. Clinical deployment requires a further step: a layer that \emph{uses} the uncertainty to decide when to answer and when to defer, that is sensitive to clinical severity, and that remains reliable when the input distribution shifts. The open question we address is therefore not how to produce a better uncertainty score in the abstract, but how to turn uncertainty into safe clinical behavior.

We develop \textit{MedBayes-Lite}, a clinical uncertainty governance layer for transformer-based clinical predictors. The layer estimates predictive uncertainty at inference time through Monte Carlo dropout, calibrates the resulting predictive distribution, and applies a confidence-guided abstention gate that defers low-confidence predictions for human review. It adds no trainable parameters and needs no retraining, so it attaches to existing pretrained encoders, and it degrades gracefully to a sampling-only mode for generative models without internal access. We deliberately evaluate it as a governance mechanism: we ask whether it prevents harmful confident errors and abstains appropriately, not whether it sets a new state of the art on uncertainty estimation.

This framing also distinguishes our study from prior uncertainty-aware transformer methods. We do not claim a new attention mechanism or a holistic layer-wise propagation result; where we incorporate uncertainty into attention scoring we treat it as one optional component of the governance layer, and our empirical claims rest on the calibration-and-abstention behavior that we can measure. This lets us report, honestly, both where the layer helps and where it does not.

Our contributions are as follows:

\begin{itemize}
    \item \textbf{A clinical uncertainty governance layer.} We recast lightweight Bayesian inference for clinical language models as a \emph{governance} problem: rather than only estimating uncertainty, the framework converts inference-time uncertainty into a risk-aware decision policy that defers low-confidence predictions. The layer combines Monte Carlo dropout uncertainty, calibration of the predictive distribution, and an entropy-based abstention gate, and attaches to pretrained encoders or sampling-only generative models without retraining or added parameters.

    \item \textbf{Reduction of harmful overconfident clinical errors.} On clinical multiple-choice QA (MedMCQA, MedQA-USMLE) the layer drives \emph{harmful} overconfident errors (confident, wrong, high clinical severity) toward zero and reduces expected calibration error by 0.23 to 0.33 absolute, with paired-bootstrap significance ($p\approx0$). The effect is strongest under domain shift, where an uncalibrated base model commits confident errors on about 21\% of high-severity items and the layer reduces this to near zero.

    \item \textbf{A comprehensive, honest clinical evaluation.} We evaluate across datasets, 13 rule-based clinical risk categories used as transparent, reproducible evaluation strata (not clinical ground-truth labels), controlled clinical perturbations, in- versus out-of-distribution transfer, and an open-weight generative contrast (Qwen2.5-7B-Instruct), reporting standard calibration and selective-prediction metrics (ECE, NLL, Brier, AURC, selective accuracy, coverage, abstention recall) alongside clinical risk metrics, and reporting negative results including a flat AURC and a null result on already-calibrated PubMedQA models.

    \item \textbf{Empirical analysis of CUS as a clinically motivated summary.} We introduce the Clinical Uncertainty Score (CUS), a severity-weighted calibration metric, and assess it empirically: across 739 model/method/category groups CUS correlates strongly with harmful overconfident error rate ($r\approx0.88$) and severity-weighted error ($r\approx0.90$). In the same analysis the Zero-shot Trustworthiness Index (ZTI) is found to be threshold-confounded and largely redundant with AURC; we therefore report CUS as a pre-specified, clinically motivated supplementary summary, while the primary selective-prediction conclusions rest on standard metrics (ECE, NLL, Brier, AURC, selective accuracy, coverage, and abstention recall); ZTI is demoted to a secondary descriptor.
\end{itemize}

The remainder of this paper is organized as follows: Section~\ref{sec:related-work} reviews existing (UQ) techniques; Section~\ref{sec:medbayes-framework} outlines the proposed framework. In Section~\ref{sec:model-settings}, we describe our dataset and provide the details of the experimental setup and our evaluation strategy. Section~\ref{sec:main-results} presents the empirical results. Section~\ref{sec:discussion} discusses broader implications and limitations. Finally, Section~\ref{sec:conclusion} concludes our paper.

\section{Related Work}
\label{sec:related-work}

The drive toward developing trustworthy clinical AI has highlighted the need for effective UQ. Accuracy alone is often insufficient in healthcare settings, where models must also signal when their predictions may be unreliable. The landscape of UQ methods for LLMs is broad, spanning foundational Bayesian techniques to clinical risk–based evaluation frameworks. However, the diversity of these approaches makes it challenging to determine which are most suitable for real-world deployment. This section synthesizes these methods to clarify how our work fits within and contributes to this space.

\subsection{Foundational Bayesian Methods in Deep Learning}

The quest to develop deep learning models that can represent their own uncertainty gained momentum with the work of Gal and Ghahramani~\cite{gal2016dropout}. They showed that dropout training can be interpreted as approximate Bayesian inference, enabling uncertainty estimation via Monte Carlo (MC) dropout without changing the underlying architecture. At inference time, MC dropout requires multiple stochastic forward passes, which can increase latency unless parallelized. MC dropout has also been applied to transformer-based models for uncertainty estimation, with some MC-dropout-based scores used to estimate epistemic uncertainty~\cite{vazhentsev2022uncertainty}. However, for clinical decision support, uncertainty is often reported only at the predictive output and does not by itself incorporate contextual or risk-aware calibration, while requiring multiple stochastic forward passes at inference can reduce efficiency and scalability.

Prior work in uncertainty quantification for deep health diagnostics has often relied on Softmax confidence or predictive entropy, but these output-level scores do not explicitly capture model uncertainty and can remain overconfident, particularly when inputs deviate from the training distribution~\cite{xia2024uncertaintyaware}. Bayesian approaches such as Monte Carlo Dropout and deep ensembles have been explored to better capture model uncertainty, but they typically increase inference-time computation due to repeated stochastic forward passes or multiple models. Evidential deep learning (EDL) offers a single-pass alternative by predicting a Dirichlet distribution over class probabilities for simultaneous prediction and uncertainty estimation. However, much of the EDL literature has been developed and evaluated on standard benchmark settings \cite{sensoy2018evidential} \cite{gao2025comprehensive}, while the behavior of uncertainty estimates under class imbalance, especially for minority classes in realistic medical data, remains less systematically studied.

\subsection{Post-Hoc Calibration and Their Limitations}

A commonly used approach to improve confidence reliability is \emph{post-hoc calibration}, which applies a probabilistic correction after training using a held-out validation set and typically assumes train/validation/test are drawn from the same distribution \cite{10.5555/3305381.3305518}. 
For multiclass models, \emph{Temperature Scaling (TS)} rescales the logit vector with a single scalar temperature parameter before softmax, adjusting confidence without changing the predicted class \cite{10.5555/3305381.3305518}. 
In contrast, \emph{Isotonic Regression (IR)} is a widely used non-parametric alternative that learns a monotone, piecewise-constant mapping from uncalibrated scores to calibrated probabilities \cite{niculescu2005predicting}. 
Because these methods are fitted on a validation set under an in-distribution protocol, their calibration quality is coupled to the calibration data and can degrade when evaluation conditions shift \cite{10.5555/3305381.3305518}. In addition to these, Ye et al. \cite{10.5555/3737916.3738407} demonstrate that output-level probability shaping, such as adjusting the softmax temperature, can substantially change the predicted probability distribution without changing the model’s accuracy. In particular, low temperatures can induce overly sharp probabilities, making the model appear overconfident even though predictive correctness is unchanged. They further show that conformal prediction is less sensitive to this effect, yielding more conservative uncertainty behavior when temperature-based probabilities become artificially confident.

\subsection{Clinical UQ and Risk-Aware Decision Making}

In healthcare, overconfident model outputs can have meaningful clinical consequences, which has motivated domain-specific work on reliability, interpretability, and human oversight. Salvi et al. \cite{salvi2025explainability} argue that Explainable AI alone is insufficient to guarantee trustworthy behavior and propose integrating Uncertainty Quantification (UQ) with XAI to improve reliability and reduce over-reliance in clinical workflows. Complementary to this direction, Catak and Kuzlu introduce a geometric UQ approach for LLMs that measures dispersion in response-embedding space via convex hull analysis, showing that uncertainty varies with prompt complexity and generation temperature \cite{catak2024uncertainty}. Together, these perspectives highlight an emerging shift from treating uncertainty as a reporting artifact toward using uncertainty signals to support cautious decision support and more robust human-AI collaboration, while fully integrating uncertainty into LLM reasoning and generation remains an open challenge.




\subsection{Benchmarking and Model Ensembles}

For uncertainty quantification benchmarking, Deep Ensembles \cite{lakshminarayanan2017simple} and Stochastic Weight Averaging Gaussian (SWAG) \cite{maddox2019simple} are widely used, high-fidelity reference methods. Deep Ensembles estimate predictive uncertainty by averaging predictions from multiple independently trained networks, which can improve robustness and calibration. However, for an ensemble of size $K$, parameter storage, model memory, training cost, and inference cost all scale approximately linearly with $K$, since the method requires maintaining and evaluating multiple independently trained models~\cite{lakshminarayanan2017simple}. SWAG instead constructs an approximate posterior over weights from the SGD trajectory and performs Bayesian model averaging by sampling multiple weight vectors at test time and averaging predictions, so inference cost increases with the number of posterior samples used~\cite{maddox2019simple}. 

In clinical informatics settings with strict latency or memory constraints, these multi-model or multi-sample procedures can be difficult to deploy at scale, motivating lightweight uncertainty estimation methods that preserve reliability and transparency while reducing ensemble-level overhead.

\subsection{Positioning MedBayes-Lite}

The current landscape of UQ for clinical LLMs reveals a clear trade-off between efficiency and the depth of uncertainty integration. Existing approaches generally fall into three categories: (i) superficial post-hoc calibration techniques that adjust model confidence without addressing internal uncertainty, (ii) computationally expensive ensemble methods that require multiple forward passes, and (iii) fragmented Bayesian implementations that apply stochasticity only to isolated components. \textit{MedBayes-Lite} is designed as a lightweight inference-time uncertainty governance layer. Rather than retraining the model or adding parameters, it injects uncertainty estimates into the inference pipeline of pretrained transformers and converts them into a calibrated, confidence-guided abstention policy. Relative to ensemble methods it adds no parameters and keeps memory close to the backbone, though its compute scales roughly linearly with the number of Monte Carlo samples. We do not claim it dominates these baselines: temperature scaling matches its calibration at lower cost and deep ensembles achieve better risk ranking. Its distinctive value is severity-aware clinical evaluation, harmful-error reduction, and improved behavior under domain shift, which we establish empirically.

\textit{MedBayes-Lite} is evaluated as a calibration-and-abstention governance layer. The empirical evaluation focuses exclusively on predictive uncertainty estimation, calibration, and confidence-guided abstention. Optional attention-level extensions are documented in Appendix~\ref{app:attention} and are not evaluated in this study. Table~\ref{tab:uq_comparison} summarizes the comparative positioning of existing UQ methods and the proposed framework.

\begin{table}[h!]
\centering
\scriptsize
\caption{Comparison of UQ Approaches for Clinical LLMs. Post-hoc, Bayesian, and ensemble methods are compared against the proposed MedBayes-Lite framework based on integration depth, computational cost, and clinical applicability.}
\label{tab:uq_comparison}
\begin{tabular}{p{0.1\textwidth}p{0.18\textwidth}p{0.2\textwidth}p{0.15\textwidth}p{0.2\textwidth}}
\hline
\textbf{Approach} & \textbf{Representative Methods} & \textbf{Integration Level} & \textbf{Comp. Overhead} & \textbf{Key Clinical Limitation} \\
\hline
Post-hoc Calibration & TS \cite{10.5555/3305381.3305518}, IR \cite{niculescu2005predicting} & Output-only & Minimal & Output-only rescaling; a strong and cheap baseline that the proposed layer matches on calibration but does not surpass. \\
\hline
Bayesian Transformers & MC Dropout on BERT \cite{joo2020being} & Partial (e.g., output/embeddings) & Low to Moderate & Uncertainty not severity-aware; no risk-aware deferral policy. \\
\hline
Ensemble Methods & Deep Ensembles \cite{lakshminarayanan2017simple}, SWAG \cite{maddox2019simple} & Model-level & Very High ($N\times$) & Computationally prohibitive for real-time deployment. \\
\hline
Proposed Framework & \textit{MedBayes-Lite} & MC-dropout + calibration + confidence-guided abstention & Memory $\approx$ backbone; compute $\propto M$ samples & Reduces harmful overconfidence and abstains under shift; matches temperature scaling on calibration and does not improve AURC. \\
\hline
\end{tabular}
\end{table}


\section{\ModelName Framework}
\label{sec:medbayes-framework}

\textit{MedBayes-Lite} is a post-training, inference-time uncertainty quantification framework for pretrained transformer models. Instead of modifying the training procedure or introducing additional learnable parameters, the framework estimates predictive uncertainty through stochastic inference and uses these uncertainty signals during inference to support calibration and confidence-guided abstention decisions. This design enables uncertainty-aware prediction without retraining the backbone architecture, which is particularly advantageous in safety-critical domains such as clinical decision support where retraining large pretrained models may be computationally or operationally impractical.

\subsection{Framework Overview}

Figure~\ref{fig:medbayes_design} illustrates the evaluated \ModelName{} pipeline used throughout this study. The evaluated framework operates through three stages. First, repeated stochastic forward passes under Monte Carlo dropout yield multiple predictions whose average forms the predictive distribution and whose dispersion provides a predictive-uncertainty signal.

Second, this predictive distribution is calibrated (for example, by temperature scaling fit on a held-out calibration set) so that the reported confidence better aligns with empirical accuracy.

Third, the predictive distribution produced by the model is converted into a normalized confidence score derived from predictive entropy. This confidence score determines whether a prediction should be accepted or deferred. Predictions with insufficient confidence can therefore be flagged for human review, enabling a risk-aware decision process.

Taken together, these stages convert inference-time uncertainty into a calibrated, confidence-guided accept-or-defer decision, without modifying the training procedure of the pretrained model.

The evaluated pipeline consists of predictive uncertainty estimation, calibration, and confidence-guided abstention. Optional attention-level extensions are described separately in Appendix~\ref{app:attention} and are not part of the evaluated framework.

\begin{figure}[ht]
    \centering
    \resizebox{\textwidth}{!}{%
    \fbox{Input} $\rightarrow$ \fbox{\shortstack{MC dropout /\\ predictive uncertainty}} $\rightarrow$ \fbox{Calibration} $\rightarrow$ \fbox{\shortstack{Confidence-guided\\ abstention}} $\rightarrow$ \fbox{Human review}}
    \caption{Evaluated \ModelName{} pipeline. Monte Carlo dropout sampling produces predictive uncertainty estimates that are calibrated and converted into a confidence-guided abstention decision. Low-confidence predictions are deferred for human review. Optional attention-level extensions are discussed separately in Appendix~\ref{app:attention} and are not part of the evaluated pipeline.}
    \label{fig:medbayes_design}
\end{figure}


\subsection{Implementation Under Different Model Access Settings}
\label{subsec:access_regimes}

The implementation of \textit{MedBayes-Lite} depends on the level of access to internal transformer states, that is, whether inference-time dropout and internal tensors are available. Accordingly, two implementation settings are considered.

\textbf{(A) Open-weight (internal-access) setting.}

For open-weight transformer models with inference-time dropout and accessible internal tensors, the evaluated method, namely MC-dropout predictive uncertainty estimation, calibration, and confidence-guided decision shaping, is implemented directly. Additional uncertainty-aware extensions are possible in this setting but are not evaluated here (Appendix~\ref{app:attention}). In this study, the experimental evaluation is restricted to encoder-based transformer models that satisfy these implementation requirements, specifically BioBERT, Bio\_ClinicalBERT, PubMedBERT, and BERT-base.

\textbf{(B) Sampling-only generative setting.}

In the sampling-only generative setting, internal embeddings, hidden states, and attention tensors are not used. Uncertainty is estimated at the output level by collecting multiple stochastic generations and constructing an empirical predictive distribution over the resulting outputs. Confidence is then computed from this distribution using entropy-based and self-consistency statistics, after which the same confidence-guided abstention rule is applied.

We evaluate a local open-weight generative model (Qwen2.5-7B-Instruct) under a restricted output-only protocol: predictive uncertainty is estimated only from sampled generations (self-consistency, semantic entropy, and a verbalized confidence). We do not claim that these results generalise to proprietary closed-API systems. Results in this configuration reflect output-level predictive uncertainty estimation rather than internal uncertainty integration.

When internal transformer states are accessible, additional uncertainty-aware extensions may be implemented. However, all reported results in this study reflect only predictive uncertainty estimation, calibration, and confidence-guided abstention. In the sampling-only generative setting, uncertainty estimation and abstention are based on observable output distributions.


\subsection{Core Components}

The framework comprises two primary components, evaluated in this paper, corresponding to the stages described in the overview: MC-dropout predictive uncertainty estimation and confidence-guided decision shaping. An optional uncertainty-weighted attention component, inactive ($\lambda=0$) in all reported experiments, is described in Appendix~\ref{app:attention} for completeness and is not evaluated here. To maintain consistent notation throughout the presentation of these components, Table~\ref{tab:notation} summarizes the symbols used in the subsequent derivations and algorithmic descriptions.

\begin{table}[h!]
\centering
\caption{Notation summary for the \textit{MedBayes-Lite} framework. Symbols and variables are defined for reference throughout Section~\ref{sec:medbayes-framework}.}
\label{tab:notation}
\renewcommand{\arraystretch}{1.2}
\setlength{\tabcolsep}{6pt}
\begin{tabular}{p{3.2cm}p{9.8cm}}
\hline
\textbf{Symbol} & \textbf{Definition} \\
\hline
$X = \{x_{1}, \ldots, x_{n}\}$ & Input clinical text sequence consisting of $n$ tokens. \\
\hline
$f_{\theta}(\cdot)$ & Transformer mapping with parameters $\theta$. \\
\hline
$z_m$ & Dropout mask (Bernoulli latent variables) sampled at inference for MC pass $m$. \\
\hline
$M$ & Number of MC dropout samples used to estimate uncertainty. \\
\hline
\multicolumn{2}{p{13cm}}{\textit{Notation for the embedding-level statistics and the optional uncertainty-weighted attention mechanism ($h^{(m)}(x),\ \hat\mu(x),\ \hat\Sigma(x),\ U(x_j),\ e_{ij},\ \tilde e_{ij},\ \alpha_{ij},\ \tilde\alpha_{ij},\ \lambda$) is provided in Appendix~\ref{app:attention}.}} \\
\hline
$p=[p_1,\ldots,p_K]$ & Predictive probability distribution over $K$ possible classes. \\
\hline
$H(p)$ & Shannon entropy: $H(p)=-\sum_{k=1}^K p_k\log p_k$. \\
\hline
$C(p)$ & Normalized confidence: $C(p)=1-\frac{H(p)}{\log K}$. \\
\hline
$\tau$ & Abstention threshold: predict only if $C(p)\ge \tau$. \\
\hline
\end{tabular}
\end{table}

\subsubsection{Predictive Uncertainty Estimation via MC Dropout}

The first stage estimates predictive uncertainty. Rather than relying on a single deterministic forward pass, \ModelName{} enables dropout at inference time and performs $M$ stochastic forward passes, averaging the resulting predictive distributions into a single predictive distribution,
\[
\bar p(x) = \frac{1}{M}\sum_{m=1}^{M} \mathrm{softmax}\!\big(f_\theta(x; z_m)\big),
\]
where $z_m$ denotes the dropout mask of pass $m$. This averaged distribution $\bar p(x)$ is the predictive distribution used by the downstream calibration and confidence-guided abstention stages, and its dispersion (for example, its predictive entropy) provides an uncertainty estimate derived from MC-dropout sampling. This follows the standard interpretation of MC dropout as approximate variational inference~\cite{gal2016dropout}. Embedding-level representation statistics and the optional uncertainty-weighted attention component that uses them are described in Appendix~\ref{app:attention} and are not part of the evaluated method.

\subsubsection{Confidence-Guided Decision Shaping}

The final stage implements the clinical principle of ``when in doubt, defer.'' For a predictive distribution
\[
p = [p_1, \ldots, p_K],
\]
the model computes a normalized confidence score
\[
C(p) = 1 - \frac{H(p)}{\log K},
\qquad
H(p) = - \sum_{k=1}^{K} p_k \log p_k .
\]

The normalization by \(\log K\) is used because \(\log K\) is the maximum possible value of the Shannon entropy over a \(K\)-class distribution, attained when \(p\) is uniform. Dividing by \(\log K\) therefore maps entropy to the interval \([0,1]\), which in turn ensures that the confidence score \(C(p)\) also lies in \([0,1]\).

A prediction is accepted only if
\[
C(p) \ge \tau,
\]
otherwise it is flagged as uncertain for human review.

Here, \(\tau \in [0,1]\) is a user-specified constant threshold that controls the abstention policy. Larger values of \(\tau\) make the decision rule more conservative by requiring greater predictive concentration before accepting a prediction.

\paragraph{Remark 3 (Interpretation of the confidence gate).}
The confidence score \(C(p)\) is a monotone transformation of the predictive entropy, with larger values corresponding to more concentrated predictive distributions. The decision rule \(C(p)\ge \tau\) therefore defines an acceptance--rejection mechanism based on predictive concentration: low-entropy outputs are accepted, whereas high-entropy outputs are deferred. In the clinical setting considered here, this rule provides a simple mechanism for abstention when the model's predictive distribution does not support a sufficiently concentrated decision.

\subsection{Algorithmic Summary}

Algorithm~\ref{alg:BLLM} summarizes the operational workflow of \ModelName. Given an input sequence, the procedure first performs repeated stochastic forward passes under MC dropout and averages them into a predictive distribution. This distribution is optionally calibrated and then mapped to an entropy-based confidence score, which determines whether the prediction is accepted or deferred according to the abstention threshold.

This algorithmic view complements the component-level formulation presented above by showing how the three stages of the framework are executed sequentially within the inference pipeline. It also clarifies that \ModelName is implemented as an inference-time procedure that reuses the pretrained transformer backbone without introducing additional trainable parameters.

\paragraph{Complexity.} The cost is dominated by the $M$ stochastic forward passes, so runtime scales as $\mathcal{O}(M\,T_{\text{fwd}})$ where $T_{\text{fwd}}$ is one forward pass; the framework adds no trainable parameters and reuses the backbone, so memory is essentially unchanged. The term \emph{Lite} refers to this parameter- and memory-efficiency: dropout-based posterior sampling rather than multiple trained models, and compatibility with existing pretrained checkpoints without retraining. Detailed latency and memory measurements are reported in Section~\ref{subsec:efficiency}.

\begin{algorithm}[ht]
\caption{High-level Pseudocode of \ModelName (evaluated pipeline)}
\label{alg:BLLM}
\KwIn{Input sequence $x=(x_1,\dots,x_n)$, MC samples $M$, optional temperature $T$, abstention threshold $\tau$}
\KwOut{Prediction $\hat y$, confidence $C(p)$}

\textbf{1) MC-Dropout Sampling and Predictive Distribution} \\
\For{$m=1$ \KwTo $M$}{
    Enable inference-time dropout (mask $z_m$) and compute $p^{(m)}=\mathrm{softmax}\!\big(f_{\theta}(x; z_m)\big)$\;
}
Form the predictive distribution (Bayesian model average) $\bar p=\frac{1}{M}\sum_{m=1}^M p^{(m)}$\;

\textbf{2) Calibration (optional)} \\
$p \leftarrow$ temperature-calibrated $\bar p$ (temperature $T$ fit on a held-out calibration set)\;

\textbf{3) Confidence-Guided Decision Shaping} \\
Compute entropy $H(p)=-\sum_{k=1}^K p_k\log p_k$ and normalized confidence $C(p)=1-\frac{H(p)}{\log K}$\;
\eIf{$C(p)\ge \tau$}{
    $\hat y=\arg\max_k p_k$\;
}{
    $\hat y=\text{``Uncertain (defer to clinician)''}$\;
}

\Return{$\hat y,\, C(p)$}\;
\end{algorithm}

\noindent\textit{Note: an optional uncertainty-weighted attention stage (Appendix~\ref{app:attention}) may be inserted before the predictive distribution is formed; it was inactive ($\lambda=0$) in all reported experiments and is therefore omitted from the evaluated pipeline.}

\section{Experimental Setup}
\label{sec:model-settings}

Having introduced the conceptual design and architectural components of \ModelName, we now describe the experimental setup used to evaluate its effectiveness. This section presents the datasets, evaluation metrics, training configurations, task protocols, and baseline methods used in our study. The objective is to ensure that all experiments are conducted under controlled, transparent, and reproducible conditions, thereby enabling a fair assessment of \ModelName\ with respect to calibration, uncertainty quantification, and clinical trustworthiness.

\subsection{Dataset Details}
\label{sec:dataset-details}

A central aspect of evaluating \ModelName\ is the selection of datasets that reflect both the linguistic complexity of biomedical text and the practical demands of clinical decision-making. To this end, we use three complementary, publicly available, de-identified clinical question-answering benchmarks: MedMCQA, MedQA-USMLE, and PubMedQA. Together they cover exam-style clinical knowledge, licensing-style diagnostic reasoning, and literature-based evidence synthesis, and they support both in-distribution evaluation and a controlled domain-shift study (Table~\ref{tab:dataset-overview}). MIMIC-III electronic health records require credentialed access and are not evaluated in this study.

\textbf{PubMedQA:} PubMedQA~\cite{jin2019pubmedqa} is a biomedical question-answering benchmark constructed from PubMed, a large-scale repository of biomedical literature. It pairs clinical questions with answers derived from research abstracts, making it suitable for \textit{literature-based question answering} and \textit{evidence synthesis for decision support}. In our experiments, we use the supervised \texttt{pqa\_labeled} split and follow the official split definitions. When controlled-size evaluation is required due to computational constraints, we report the exact subset sizes used in Table~\ref{tab:dataset-overview}.

\textbf{MedQA:} MedQA~\cite{yang2025llmmedqaenhancingmedicalquestion} consists of clinically relevant multiple-choice questions curated from medical licensing-style examinations and standard medical references. It reflects common diagnostic and treatment questions encountered in medical reasoning and is therefore well aligned with \textit{diagnostic reasoning} and \textit{clinical decision-making} tasks. We follow the official dataset partitions, and any reduced subset used for computational efficiency is explicitly reported in Table~\ref{tab:dataset-overview}.

\textbf{MedMCQA:} MedMCQA is a large-scale multiple-choice clinical question-answering benchmark drawn from medical entrance-examination questions spanning many clinical subjects. It is our primary in-distribution multiple-choice task ($n=3{,}683$ test items) and the source domain for the domain-shift experiment, in which MedMCQA-trained models are evaluated out-of-distribution on MedQA-USMLE. Exact split sizes are reported in Table~\ref{tab:dataset-overview}.

\begin{table}[h!]
\centering
\caption{Datasets used in the experimental evaluation. All three are publicly available
de-identified biomedical QA benchmarks. MIMIC-III electronic health records require
credentialed access and were not evaluated in this study; the in-hospital mortality task is
left as a separable future axis (a credentialed-access adapter is provided in the code).
Reported sizes correspond to the exact test splits used.}
\label{tab:dataset-overview}
{\scriptsize
\begin{tabular}{p{2.4cm}p{3.0cm}p{1.6cm}p{5.4cm}}
\hline
\textbf{Dataset} & \textbf{Task} & \textbf{Test size} & \textbf{Role in this study} \\
\hline
MedMCQA & 4-way single-best-answer clinical exam QA & 3{,}683 & Primary in-distribution
multiple-choice task; source domain for the domain-shift experiment. \\
\hline
MedQA-USMLE & 4-way clinical licensing-exam QA & 973 & Out-of-distribution target for
MedMCQA-trained models; also evaluated directly. \\
\hline
PubMedQA & 3-way (yes/no/maybe) evidence QA & 350 & Literature-based QA; used with a locally
fine-tuned encoder family that is already well calibrated (a null-result control). \\
\hline
\end{tabular}
}
\end{table}

\subsection{Clinical Risk Category Assignment}
\label{subsec:risk-categories}

To evaluate clinical safety beyond aggregate accuracy, we stratify every item into one or more rule-based \emph{clinical risk categories}. Each category is assigned by transparent keyword and regular-expression rules over the item text (for example, drug-name suffixes and dose patterns for the medication category, or a count of negation cues for the negation-heavy category), so an item may belong to several categories at once. Each category carries a severity weight encoding the clinical cost of a confident error; these weights define harmful overconfident errors (those in high-severity categories, severity $\geq 2.5$) and weight the Clinical Uncertainty Score. Table~\ref{tab:risk_categories} lists the categories, example rules, severity weights, and clinical motivation. These categories are heuristic evaluation strata, not ground-truth clinical labels; they give a transparent, reproducible way to ask where a model's confident errors fall, and are hereafter used as evaluation strata.

\begin{table}[h]
\centering
\caption{Rule-based clinical risk categories used as evaluation strata. Each item is assigned to
one or more categories by transparent keyword and regular-expression rules over its text; the
severity weight is used to define harmful overconfident errors (severity $\geq 2.5$) and to weight
the Clinical Uncertainty Score. These categories are heuristic evaluation strata, not
ground-truth clinical labels.}
\label{tab:risk_categories}
\renewcommand{\arraystretch}{1.2}
\scriptsize
\begin{tabular}{p{2.7cm}p{5.3cm}p{1.1cm}p{3.4cm}}
\hline
\textbf{Clinical risk category} & \textbf{Rule examples} & \textbf{Sev.\ wt.} & \textbf{Clinical motivation} \\
\hline
Medication / drug interaction & drug-name suffixes (-pril, -olol, -statin, -mab); dose amounts (e.g. 500\,mg); ``interaction'', ``contraindicated'' & 3.0 & Dosing and interaction errors translate directly into patient harm. \\
Diagnosis / differential & ``diagnosis'', ``differential'', ``most likely'', ``consistent with'' & 3.0 & Misdiagnosis drives downstream management errors. \\
Treatment recommendation & ``treat'', ``therapy'', ``management'', ``first-line'', ``next step'' & 3.0 & An incorrect treatment recommendation is directly actionable and harmful. \\
Risk / prognosis & ``prognosis'', ``mortality'', ``survival'', ``recurrence'' & 2.5 & Miscommunicated risk distorts shared decision-making. \\
Rare / uncommon condition & ``rare'', ``atypical'', ``syndrome'', ``congenital'' & 2.5 & Rare presentations are high-uncertainty and high-stakes. \\
Laboratory / biomarker & ``serum'', ``creatinine'', ``WBC'', ``mg/dl'', ``titer'' & 2.0 & Lab misinterpretation propagates to diagnosis and dosing. \\
Numeric / vital-heavy & $\geq 4$ numeric or unit tokens (mmHg, bpm, mg/dL) & 2.0 & Numeric-dense items are prone to value misreading. \\
Ambiguous symptoms & ``nonspecific'', ``vague'', ``could be'', ``atypical presentation'' & 2.0 & Ambiguity warrants more conservative deferral. \\
Missing / incomplete evidence & ``not available'', ``unknown'', ``insufficient'', ``no history'' & 2.0 & Incomplete evidence should lower confidence. \\
Contradictory evidence & ``however'', ``despite'', ``inconsistent'', ``conflicting'' & 2.0 & Conflicting findings require cautious aggregation. \\
Negation-heavy text & $\geq 3$ negation cues (no, not, denies, absent) & 1.5 & Negation flips clinical meaning and is error-prone. \\
High-severity case & ``acute'', ``severe'', ``critical'', ``ICU'', ``sepsis'', ``STAT'' & 3.0 & Emergent cases carry the highest cost of confident error. \\
Low-severity (routine) & fallback when no higher-risk category fires & 1.0 & Routine factual QA; baseline severity. \\
\hline
\end{tabular}
\end{table}

\subsection{Evaluation Metrics}
\label{app:evaluation_metrics}

To evaluate the calibration and clinical reliability of \ModelName, we employ both standard uncertainty quantification metrics and two clinically motivated measures. This combination allows the models to be assessed not only in terms of statistical calibration, but also in terms of properties relevant to safe clinical deployment.

\subsubsection{Standard Calibration Metrics}

\begin{itemize}
    \item \textbf{Expected Calibration Error (ECE)}~\cite{10.5555/3305381.3305518}: ECE measures the discrepancy between model confidence and empirical accuracy. Predictions are partitioned into $B=15$ confidence bins, and ECE is computed as
    \[
    \text{ECE} = \sum_{b=1}^{B} \frac{|B_b|}{N} \left|\text{acc}(B_b) - \text{conf}(B_b)\right|,
    \]
    where $\text{acc}(B_b)$ and $\text{conf}(B_b)$ denote the accuracy and average confidence within bin $B_b$, respectively. Lower ECE indicates better calibration.
    
    \item \textbf{Negative Log-Likelihood (NLL)}: NLL is a proper scoring rule that penalizes inaccurate and poorly calibrated probabilistic predictions. For a classification task with $K$ classes,
    \[
    \text{NLL} = -\frac{1}{N}\sum_{i=1}^{N} \sum_{k=1}^{K} y_{i,k} \log(p_{i,k}),
    \]
    where $y_{i,k}$ is the ground-truth indicator and $p_{i,k}$ is the predicted probability for class $k$. Lower NLL indicates better probabilistic predictions.
\end{itemize}

\subsubsection{Clinically Motivated Trustworthiness Metrics}

\begin{itemize}
    \item \textbf{CUS}: We introduce a weighted calibration metric that accounts for clinical risk by assigning larger penalties to overconfident predictions in safety-critical scenarios. Unlike standard ECE, CUS incorporates clinical severity weights $w_{\text{clinical}}(b)$ that scale with the potential harm associated with an error:
    \[
    \text{CUS} = \sum_{b=1}^{B} \left|\text{acc}(b) - \text{conf}(b)\right| \times P(b) \times w_{\text{clinical}}(b),
    \]
    where $b$ indexes confidence bins, $\text{acc}(b)$ and $\text{conf}(b)$ denote the accuracy and mean confidence in bin $b$, $P(b)$ is the proportion of samples in bin $b$, and $w_{\text{clinical}}(b)$ assigns larger weight to high-confidence errors in clinically severe cases. Lower CUS values indicate more clinically aware uncertainty estimation. We empirically evaluate CUS in Section~\ref{sec:main-results}: across model/method/category groups it correlates strongly with the harmful overconfident error rate ($r\approx0.88$), so it is reported as a pre-specified, clinically motivated supplementary summary. The primary selective-prediction conclusions are based on standard metrics: ECE, NLL, Brier score, AURC, selective accuracy, coverage, and abstention recall.
    
    \item \textbf{ZTI}: This metric formalizes the trade-off between \emph{coverage} (the fraction of instances for which the model makes a prediction) and \emph{reliability} (the accuracy of predictions when the model is sufficiently confident). ZTI rewards models that maintain strong predictive performance while abstaining when uncertainty is high:
    \[
    \text{ZTI} = 2 \times \frac{\text{Coverage} \times \text{Calibrated Accuracy}}{\text{Coverage} + \text{Calibrated Accuracy}},
    \]
    where
    \[
    \text{Coverage} = \frac{|\mathcal{S}_{\text{confident}}|}{|\mathcal{S}_{\text{total}}|}, \quad 
    \text{Calibrated Accuracy} = \text{Accuracy on } \mathcal{S}_{\text{confident}}.
    \]
    Here, $\mathcal{S}_{\text{confident}}$ denotes the set of predictions whose confidence exceeds a threshold $\tau$, and $\mathcal{S}_{\text{total}}$ denotes the full test set. Higher ZTI values nominally indicate a better balance between applicability and safety. However, we find ZTI to be threshold-confounded and largely redundant with the area under the risk-coverage curve (Section~\ref{sec:main-results}); we therefore report it only as a secondary descriptor and rely on AURC, selective accuracy, coverage, and abstention recall for selective-prediction claims.
\end{itemize}

\subsubsection{Metric Alignment with Clinical Objectives}

The relationship between these metrics and the goals of \ModelName\ is deliberate. Calibration of the predictive distribution reduces overconfidence, thereby improving both ECE and CUS. The confidence-guided abstention component governs the trade-off between coverage and accepted accuracy, which we report primarily through AURC, selective accuracy, coverage, and abstention recall; ZTI is reported only as a secondary descriptor. Taken together, these metrics provide a broader assessment of whether a model is not only accurate, but also sufficiently reliable for clinical use. Their roles and interpretations are summarized in Table~\ref{tab:metrics_summary}.

\begin{table}[h]
\centering
\caption{Summary of evaluation metrics and their clinical interpretation. Primary metrics are ECE, NLL, AURC, selective accuracy, coverage, and abstention recall. CUS is reported as a clinically motivated supplementary calibration summary. ZTI is reported for completeness but is not used for primary conclusions, because subsequent analysis (Section~\ref{sec:main-results}) shows it is threshold-confounded and largely redundant with AURC.}
\label{tab:metrics_summary}
\begin{tabular}{p{1.5cm}p{10cm}p{3cm}}
\hline
\textbf{Metric} & \textbf{Clinical Interpretation} & \textbf{Desired Direction} \\
\hline
ECE & Alignment between predictive confidence and empirical accuracy & $\downarrow$ (Lower is better) \\
\hline
NLL & Quality of probabilistic predictions & $\downarrow$ (Lower is better) \\
\hline
CUS & Clinically motivated \textit{supplementary} calibration summary (severity-weighted calibration error) & $\downarrow$ (Lower is better) \\
\hline
ZTI & Secondary descriptor (coverage/accuracy harmonic mean; threshold-confounded) & $\uparrow$ (Higher is better) \\
\hline
\end{tabular}
\end{table}

\subsection{Training Configuration}
\label{sec:training_config}

To ensure consistency and fairness in evaluation, all models were trained and evaluated under standardized configurations. Table~\ref{tab:model_settings} summarizes the key hyperparameters used across model families, including sequence length, temperature, and token limits. These settings were selected to balance reproducibility, computational efficiency, and the need to capture extended clinical context. The maximum sequence length was fixed at 512 tokens for all models to support long biomedical narratives, such as PubMed abstracts and MIMIC-III clinical notes, while preserving computational feasibility.

\textbf{Fine-tuning of discriminative models.}  
For BERT-based architectures (BERT, BioBERT, ClinicalBERT, and RoBERTa), we used the \textit{AdamW} optimizer with learning rates between $2\times10^{-5}$ and $5\times10^{-5}$, batch sizes of 16--32, and 3--5 training epochs depending on convergence. Early stopping was applied based on validation loss, with checkpoints saved when validation loss did not improve for three consecutive epochs. A validation split was reserved from each dataset, and metrics such as ECE and NLL were used to guide model selection. These settings follow established fine-tuning practice for transformer-based biomedical classification tasks.

\textbf{Generative model configuration.}  
For the open-weight generative model (Qwen2.5-7B-Instruct) we use a sampling-only mode: predictive uncertainty is estimated from multiple stochastic generations (self-consistency and semantic entropy) and from an elicited verbalized confidence, after which the same confidence-guided abstention rule is applied. Generative outputs are constrained to short answers for PubMedQA (Yes/No/Maybe) and to a single option letter for the multiple-choice tasks, as described in Section~\ref{subsec:qa_eval_protocol}. Proprietary GPT-class API models, GPT-2, and LLaMA-2 were not evaluated.

\textbf{Bayesian dropout and uncertainty estimation.}  
MC dropout is used as an inference-time uncertainty estimator within \ModelName{} via repeated stochastic forward passes. Unless otherwise stated, all reported experiments use an inference dropout rate of $0.2$ and $15$ Monte Carlo samples. A sensitivity analysis over dropout rates $0.1$--$0.5$ and Monte Carlo sample counts $5$--$50$ is reported in the Supplementary Material (Tables~\ref{tab:sens_dropout} and~\ref{tab:sens_mc}): calibration and harmful-error metrics are stable with respect to the number of Monte Carlo samples (for $M\ge10$), and increasing the dropout rate trades improved calibration for reduced coverage and risk ranking, with $0.2$ providing a balanced operating point. Only calibration and confidence-guided abstention are evaluated in this study; the optional attention pathway remained inactive ($\lambda=0$).

\textbf{Validation and model selection.}  
Each model is validated using stratified splits to preserve representative coverage of common and rare clinical cases. Early stopping based on validation loss, together with repeated evaluation across random seeds, is used to improve robustness and reproducibility.

\textbf{Temperature scaling for discriminative models.}  
Although temperature is typically associated with generative decoding, we also apply temperature scaling to discriminative architectures as a post-hoc calibration method at the logit level. Temperature scaling reduces overconfidence by rescaling logits prior to softmax normalization, thereby improving the alignment between predicted probabilities and empirical accuracy in safety-sensitive settings.

Overall, the configurations in Table~\ref{tab:model_settings} define a uniform experimental protocol across model families. This controlled setup enables reliable comparison of the uncertainty quantification and trustworthiness measures used to evaluate \ModelName.

\begin{table}[h!]
\centering
\caption{Models evaluated in this study. Encoder predictors are evaluated with full internal
access (Monte Carlo dropout, calibration, abstention); the generative model is evaluated in a
sampling-only generative mode. We report three random seeds and 15 Monte Carlo dropout
samples (inference dropout rate 0.2). GPT-class API models, GPT-2, and LLaMA-2 were not run.}
\label{tab:model_settings}
\renewcommand{\arraystretch}{1.2}
\setlength{\tabcolsep}{8pt}
\begin{tabular}{p{5.6cm}cc}
\hline
\textbf{Model} & \textbf{Task / head} & \textbf{Max length} \\
\hline
BioBERT (MedMCQA-tuned) & 4-way multiple choice & 320 \\
PubMedBERT (MedMCQA-tuned) & 4-way multiple choice & 320 \\
BERT-base (fine-tuned) & PubMedQA 3-way & 512 \\
BioBERT (fine-tuned) & PubMedQA 3-way & 512 \\
Bio\_ClinicalBERT (fine-tuned) & PubMedQA 3-way & 512 \\
PubMedBERT (fine-tuned) & PubMedQA 3-way & 512 \\
Qwen2.5-7B-Instruct (generative) & sampling-only (self-consistency / & 512 \\
 & semantic entropy / verbalized) & \\
\hline
\end{tabular}
\end{table}

\subsection{Task Protocols and Output Standardization}
\label{subsec:qa_eval_protocol}

Because the suite includes both discriminative encoder classifiers (BioBERT, Bio\_ClinicalBERT, PubMedBERT, BERT-base) and a generative large language model (Qwen2.5-7B-Instruct), we standardize the evaluation protocol to ensure comparable task-level outputs across model types.

\textbf{PubMedQA (3-way classification).}
For encoder models, we fine-tune a three-class classifier head and evaluate the predicted labels directly. For generative models, we prompt the model to output exactly one label from the set \{\texttt{Yes}, \texttt{No}, \texttt{Maybe}\}. Outputs outside this set are mapped to the nearest valid label using a deterministic normalization rule based on case-insensitive string matching; otherwise, they are marked invalid and excluded from accuracy while still included in abstention statistics.

\textbf{MedQA (multiple choice).}
For encoder models, we fine-tune a 4-way (or 5-way, when applicable) classifier head. For generative models, we prompt the model to output exactly one option letter (e.g., \texttt{A/B/C/D}). If a model produces additional text, only the first valid option letter is retained. If no valid option is found, the sample is treated as invalid and counted as an abstention for coverage calculations.

\textbf{MedMCQA (multiple choice).}
MedMCQA is evaluated identically to MedQA-USMLE: encoder models score the four options and the generative model outputs a single option letter, under the same normalization and abstention handling.

\textbf{Uncertainty estimation and abstention.}
For open-weight models, uncertainty signals are computed from inference-time stochasticity under MC dropout and used for confidence-guided decision shaping; the optional attention pathway was inactive ($\lambda=0$). In a sampling-only generative setting, predictive uncertainty may be estimated from repeated sampled generations, after which the same confidence-guided abstention rule can be applied. We report both \emph{accuracy on accepted predictions} and \emph{coverage} (the fraction of non-abstained outputs), which together determine ZTI.

\subsection{Baseline Implementations}
\label{app:baseline-implementations}

For comparison, we implement several widely used post-hoc calibration baselines alongside \ModelName. Maximum Softmax Probability (MSP) uses the maximum predicted class probability as a confidence proxy~\cite{hendrycks2018baselinedetectingmisclassifiedoutofdistribution}. Temperature Scaling (TS) applies a scalar temperature parameter, optimized on the validation set using LBFGS to minimize cross-entropy loss, thereby rescaling logits prior to softmax. Isotonic Regression (IR) provides a non-parametric mapping between predicted confidence and empirical accuracy and is implemented using scikit-learn’s monotone regression with clipping for out-of-range predictions~\cite{niculescu2005predicting}. All baselines are applied to the same predicted logits obtained from the Bio\_ClinicalBERT backbone and are evaluated using ECE, NLL, CUS, and ZTI. For reproducibility, ECE is computed with 15 bins, TS optimization uses a learning rate of 0.01 for 50 iterations, and IR is trained on raw top-class confidence values.

These baselines represent commonly used calibration strategies and provide reference points against which \ModelName\ is compared. For the locally fine-tuned encoder family, all baselines start from the same pretrained backbone and use identical data splits, preprocessing, and optimization, so that differences are attributable to the uncertainty-handling strategy rather than to initialization or data. Deep ensembles are formed from independently fine-tuned seeds; as reported in Section~\ref{sec:main-results}, they achieve the best risk ranking among the methods compared. We use 15 Monte Carlo dropout samples at inference (dropout rate 0.2) throughout.

To avoid ambiguity regarding what is implementable under different access settings, we explicitly summarize which \textit{MedBayes-Lite} components apply to open-weight models and which apply in the sampling-only generative setting. Table~\ref{tab:component_applicability} provides this mapping and clarifies that the optional embedding- and attention-level extensions require internal access and were not evaluated in this study, whereas confidence-guided abstention applies in both settings.

\begin{table}[h!]
\centering
\caption{Applicability of \textit{MedBayes-Lite} components under different model access settings. The open-weight (internal-access) mode permits inference-time dropout and internal interventions, whereas the sampling-only generative mode is limited to output-level sampling-based uncertainty estimation and confidence-guided abstention. The confidence-guided abstention used throughout this study applies in both settings.}
\label{tab:component_applicability}
\renewcommand{\arraystretch}{1.15}
\small
\begin{tabular}{p{3.2cm}|c|c}
\hline
\textbf{Component} & \textbf{Open-weight models} & \textbf{Sampling-only generative setting} \\
\hline
Predictive Uncertainty Estimation (MC-dropout sampling) & \checkmark & -- \\
\hline
Uncertainty-Weighted Attention (optional; inactive in reported experiments) & \checkmark & -- \\
\hline
Confidence-Guided Decision Shaping (abstention) & \checkmark & \checkmark \\
\hline
Predictive uncertainty via multi-sample outputs (entropy / self-consistency) & optional & \checkmark \\
\hline
\end{tabular}
\end{table}

\section{Results}
\label{sec:main-results}

We evaluate \ModelName as a clinical uncertainty governance layer. The questions are whether it
reduces harmful overconfident clinical errors (Section~\ref{subsec:harmful}), improves
calibration (Section~\ref{subsec:calib}), supports appropriate abstention
(Section~\ref{subsec:abstain}), remains useful under medical domain shift
(Section~\ref{subsec:shift}), resolves to clinically meaningful risk categories
(Section~\ref{subsec:category-analysis}), and whether the proposed CUS summary is meaningful
(Section~\ref{subsec:cusval}). Section~\ref{subsec:negative} reports the negative findings that
bound these claims. All numbers are from runs over three seeds on the full encoder test splits,
with 15 Monte Carlo dropout samples; confidence intervals use a per-sample bootstrap.

\subsection{Harmful Overconfident Error Reduction}
\label{subsec:harmful}

Table~\ref{tab:main_results} reports the central result on clinical multiple-choice QA. A
\emph{harmful overconfident error} is a prediction that is incorrect, accepted at a confidence
threshold (here $0.8$), and clinically high-severity. Where the base model is overconfident,
\ModelName drives these errors toward zero and sharply reduces calibration error. On
MedMCQA with PubMedBERT, expected calibration error falls from $0.255$ to $0.023$ and the
harmful overconfident error rate from $0.056$ to $0.003$; on BioBERT, from $0.317$ to $0.021$
and $0.067$ to $0.000$. A paired bootstrap ($n=2000$) confirms significance: across the four
multiple-choice settings the harmful-error reduction ranges from $0.05$ to $0.22$ and the ECE
reduction from $0.23$ to $0.33$, all with $p\approx0$. Temperature scaling already captures
most of the calibration gain, which we discuss in Section~\ref{subsec:calib}; the mechanism of
the harmful-error reduction is calibration together with confidence-guided abstention rather
than improved error ranking (Section~\ref{subsec:negative}).

\begin{table}[h!]
\centering
\caption{Harmful overconfident error reduction on clinical multiple-choice QA. ECE and
harmful@0.8 (confident, wrong, high-severity) are lower-is-better; selective accuracy at
$\tau{=}0.7$ is higher-is-better; AURC is the area under the risk-coverage curve
(lower-is-better). \ModelName removes nearly all harmful overconfident errors and improves
calibration, but does not improve AURC.}
\label{tab:main_results}
\renewcommand{\arraystretch}{1.15}
\small
\begin{tabular}{llccccc}
\hline
\textbf{Dataset} & \textbf{Method} & \textbf{ECE}$\downarrow$ & \textbf{harmful@0.8}$\downarrow$ & \textbf{SelAcc@0.7}$\uparrow$ & \textbf{Cov@0.7} & \textbf{AURC}$\downarrow$ \\
\hline
\multicolumn{7}{l}{\textit{MedMCQA, PubMedBERT}} \\
 & base & 0.255 & 0.056 & 0.571 & 0.541 & 0.376 \\
 & temperature scaling & 0.081 & 0.012 & 0.721 & 0.217 & 0.375 \\
 & MC-dropout & 0.087 & 0.009 & 0.723 & 0.200 & 0.392 \\
 & \textit{MedBayes-Lite} & \textbf{0.023} & \textbf{0.003} & \textbf{0.818} & 0.100 & 0.392 \\
\hline
\multicolumn{7}{l}{\textit{MedMCQA, BioBERT}} \\
 & base & 0.317 & 0.067 & 0.457 & 0.508 & 0.517 \\
 & temperature scaling & 0.085 & 0.003 & 0.618 & 0.083 & 0.515 \\
 & MC-dropout & 0.163 & 0.013 & 0.541 & 0.189 & 0.521 \\
 & \textit{MedBayes-Lite} & \textbf{0.021} & \textbf{0.000} & \textbf{0.714} & 0.021 & 0.521 \\
\hline
\multicolumn{7}{l}{\textit{MedQA-USMLE, PubMedBERT}} \\
 & base & 0.325 & 0.213 & 0.437 & 0.541 & 0.539 \\
 & temperature scaling & 0.094 & 0.015 & 0.583 & 0.106 & 0.533 \\
 & MC-dropout & 0.171 & 0.045 & 0.483 & 0.201 & 0.558 \\
 & \textit{MedBayes-Lite} & \textbf{0.046} & \textbf{0.002} & 0.539 & 0.026 & 0.554 \\
\hline
\multicolumn{7}{l}{\textit{MedQA-USMLE, BioBERT}} \\
 & base & 0.410 & 0.217 & 0.288 & 0.486 & 0.724 \\
 & temperature scaling & 0.162 & 0.004 & 0.182 & 0.034 & 0.726 \\
 & MC-dropout & 0.253 & 0.051 & 0.292 & 0.164 & 0.724 \\
 & \textit{MedBayes-Lite} & \textbf{0.084} & \textbf{0.000} & 0.143 & 0.002 & 0.720 \\
\hline
\end{tabular}
\end{table}

\begin{figure}[h]
\centering
\includegraphics[width=0.8\textwidth]{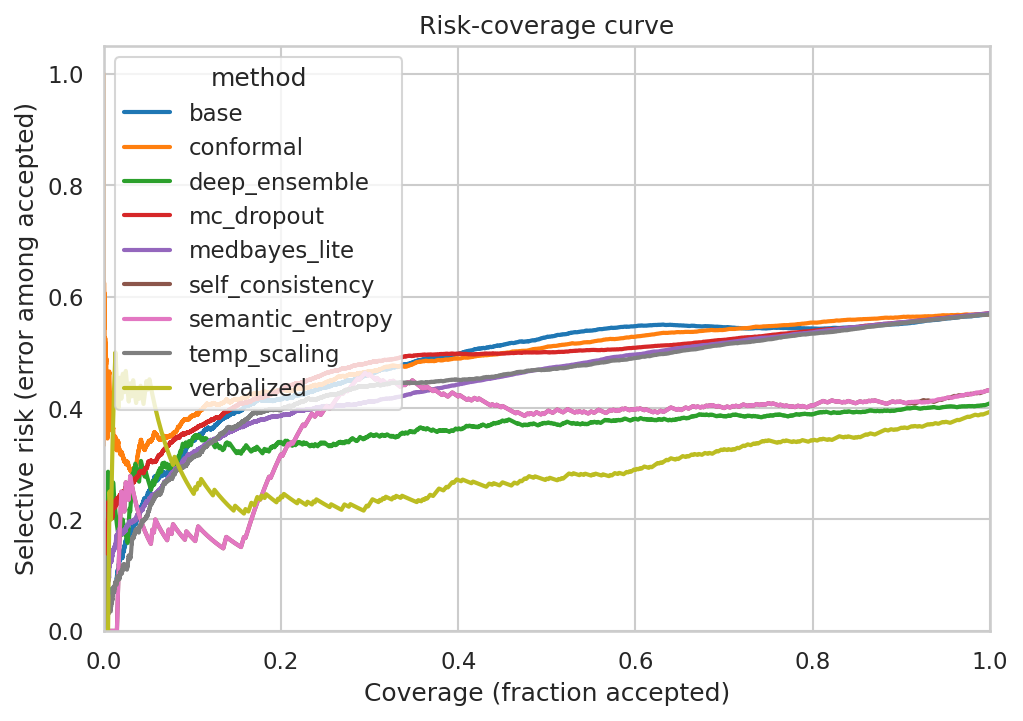}
\caption{Risk-coverage curves on MedMCQA. \ModelName and the baselines trace nearly the same
curve, illustrating that the method changes the confidence scale (and thus calibration and
threshold-based harmful-error rates) but not the underlying ordering of predictions by
reliability.}
\label{fig:risk_coverage}
\end{figure}

\subsection{Calibration Performance}
\label{subsec:calib}

\begin{figure}[h]
\centering
\includegraphics[width=0.8\textwidth]{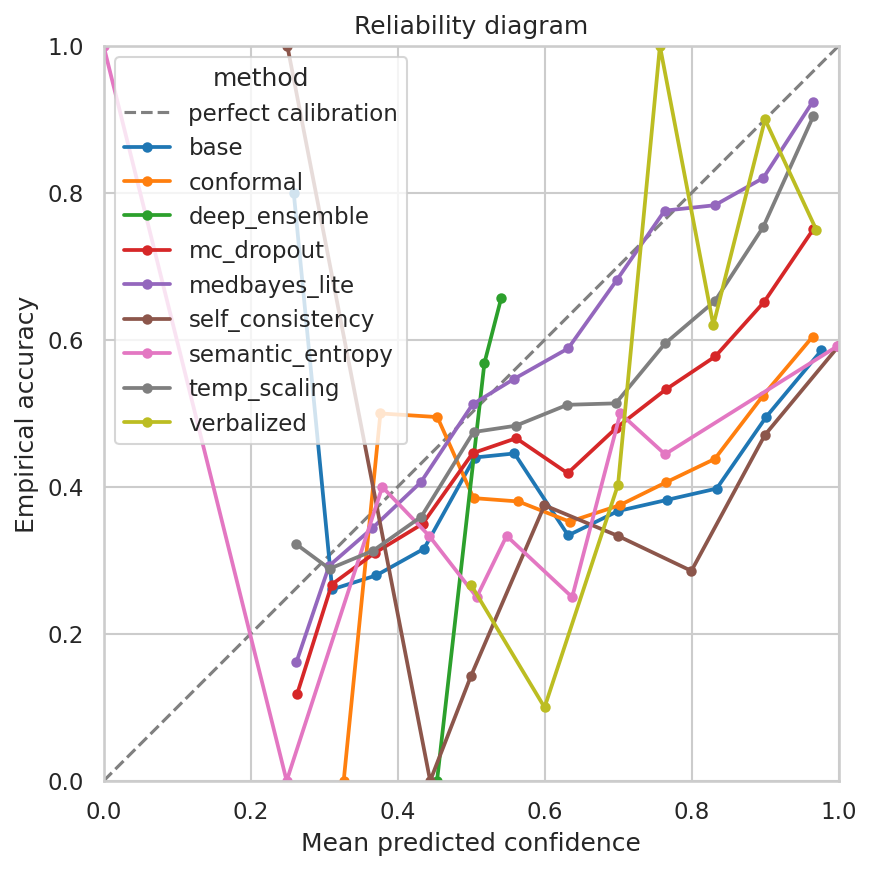}
\caption{Reliability diagrams on MedMCQA. The base model sits well below the diagonal
(overconfident); calibration and \ModelName bring confidence into agreement with accuracy.}
\label{fig:reliability}
\end{figure}

Figure~\ref{fig:reliability} shows reliability diagrams for MedMCQA. The base model is markedly
overconfident; both temperature scaling and \ModelName restore close agreement between
confidence and accuracy. As Table~\ref{tab:main_results} makes explicit, temperature scaling, a
single-parameter post-hoc method, already achieves most of the calibration gain (ECE $0.255 \to
0.081$ on PubMedBERT), and \ModelName, which adds Monte Carlo averaging on top of calibration,
reaches the lowest ECE ($0.023$) and Brier score ($0.652$ versus $0.750$ for the base model).
We therefore present temperature scaling as a strong, cheap baseline rather than a strawman.

\subsection{Risk-Aware Abstention}
\label{subsec:abstain}

\begin{figure}[h]
\centering
\begin{subfigure}{0.48\textwidth}
\centering
\includegraphics[width=\textwidth]{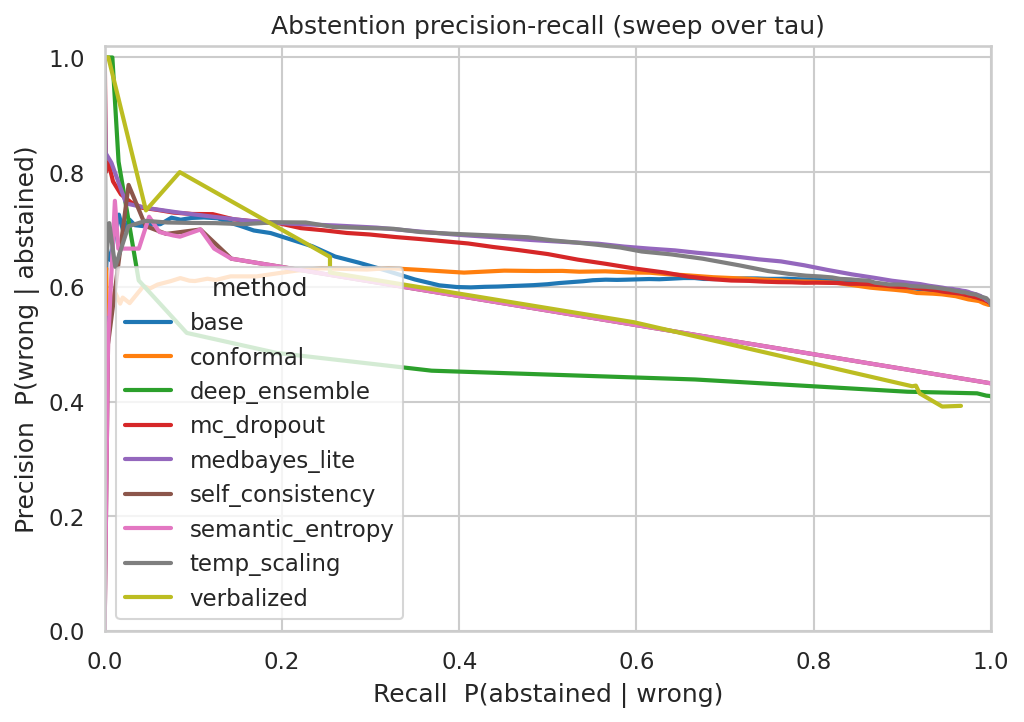}
\caption{Abstention precision-recall.}
\label{fig:abstention}
\end{subfigure}\hfill
\begin{subfigure}{0.48\textwidth}
\centering
\includegraphics[width=\textwidth]{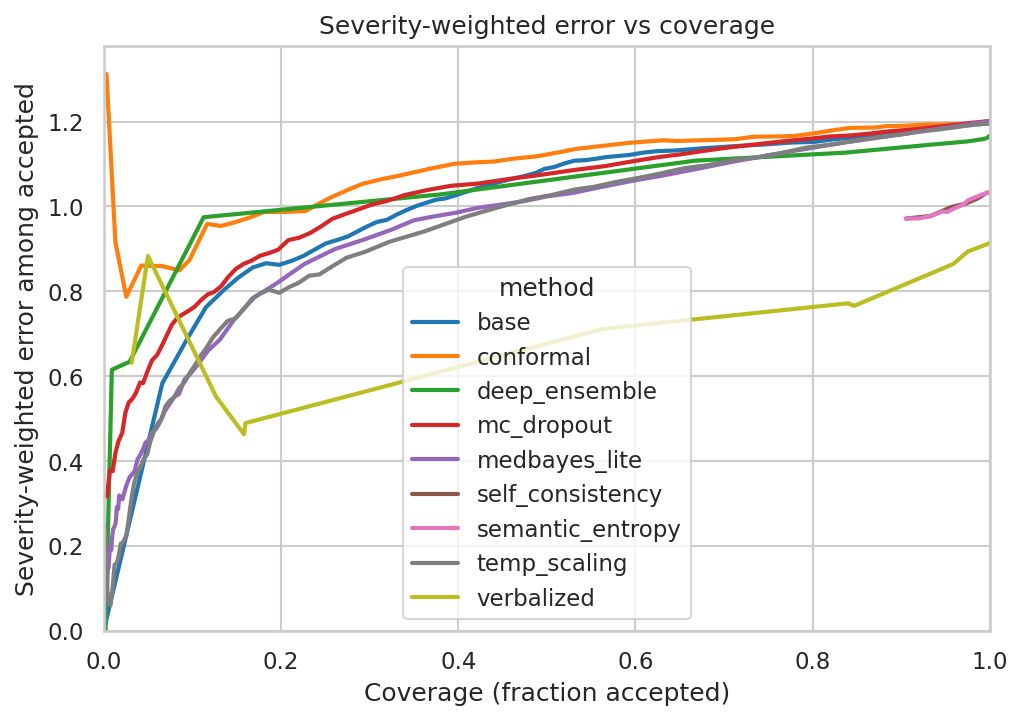}
\caption{Severity-weighted error vs coverage.}
\label{fig:sevcov}
\end{subfigure}
\caption{Selective-prediction behavior on MedMCQA. \ModelName defers a large fraction of
incorrect predictions (high abstention recall) and reduces severity-weighted error among
accepted predictions, at the cost of lower coverage.}
\end{figure}

The governance value of the layer is in how it defers. Under \ModelName the accuracy of
\emph{accepted} predictions at $\tau{=}0.7$ rises sharply (Table~\ref{tab:main_results}): from
$0.571$ to $0.818$ on MedMCQA with PubMedBERT, and from $0.457$ to $0.714$ on BioBERT.
Abstention recall, the fraction of incorrect predictions that are deferred, reaches $0.97$ to
$0.99$ at $\tau{=}0.7$ versus roughly $0.55$ for the base model. This safety is bought at a
coverage cost that we report explicitly: coverage at $\tau{=}0.7$ falls from $0.541$ to $0.100$
on MedMCQA with PubMedBERT. The model answers less often, but is right far more often when it
does. Figures~\ref{fig:abstention} and~\ref{fig:sevcov} show the abstention precision-recall
behavior and the severity-weighted error as a function of coverage.

\subsection{Domain-Shift Robustness}
\label{subsec:shift}

The strongest result is under medical domain shift. We train (or load) MedMCQA models and
evaluate them out-of-distribution on MedQA-USMLE. Table~\ref{tab:ood} reports the drift. The
base models become acutely dangerous under shift: they commit confident errors on about $21\%$
of high-severity items (harmful@0.8 of $0.213$ for PubMedBERT and $0.217$ for BioBERT).
\ModelName reduces this to near zero ($0.002$ and $0.000$) and roughly halves the calibration
drift on PubMedBERT (ECE drift $+0.070 \to +0.026$). A clinical model that is most overconfident
exactly when it is most likely to be wrong is the failure this layer is designed to contain.

\begin{table}[h]
\centering
\caption{Out-of-distribution transfer (train MedMCQA, evaluate MedQA-USMLE). ``Drift'' is the
in-distribution to out-of-distribution change in ECE. Under shift the base model commits
confident errors on $\sim$21\% of high-severity items; \ModelName reduces this to near zero.}
\label{tab:ood}
\renewcommand{\arraystretch}{1.2}
\small
\begin{tabular}{llccccc}
\hline
\textbf{Model} & \textbf{Method} & \textbf{ID ECE} & \textbf{OOD ECE} & \textbf{ECE drift} & \textbf{ID harmful} & \textbf{OOD harmful} \\
\hline
PubMedBERT & base & 0.255 & 0.325 & $+0.070$ & 0.056 & 0.213 \\
PubMedBERT & \textit{MedBayes-Lite} & 0.030 & 0.055 & $\mathbf{+0.026}$ & 0.003 & \textbf{0.002} \\
BioBERT & base & 0.317 & 0.410 & $+0.093$ & 0.067 & 0.217 \\
BioBERT & \textit{MedBayes-Lite} & 0.015 & 0.084 & $+0.069$ & 0.000 & \textbf{0.000} \\
\hline
\end{tabular}
\end{table}

\begin{figure}[h]
\centering
\includegraphics[width=0.7\textwidth]{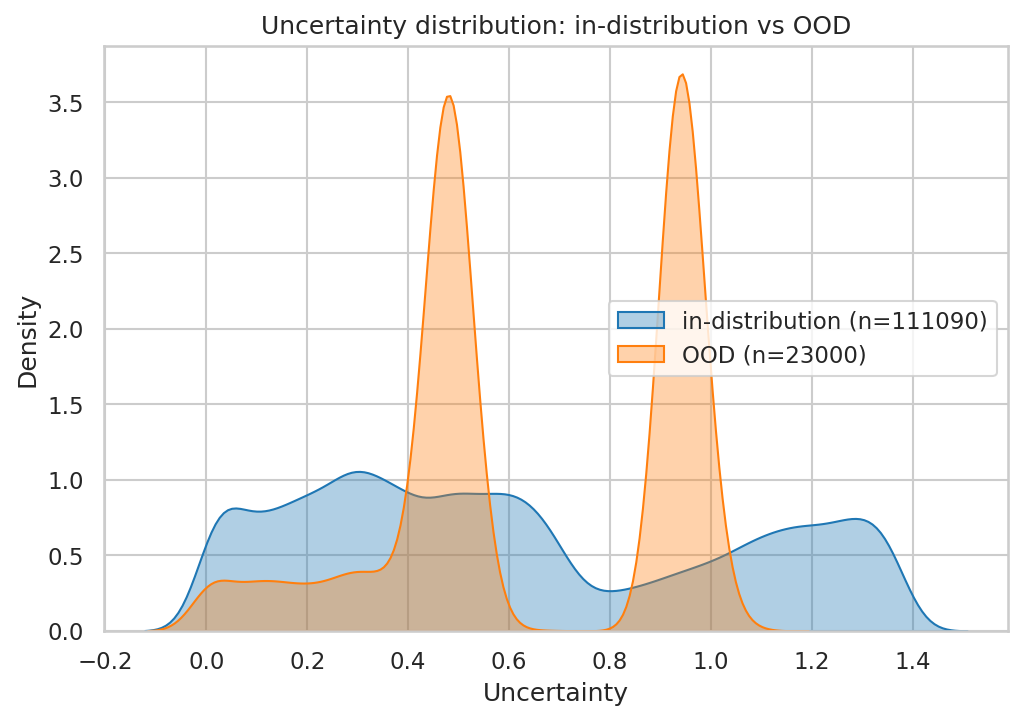}
\caption{Predictive uncertainty in-distribution (MedMCQA) versus out-of-distribution
(MedQA-USMLE). Uncertainty shifts upward under domain shift, which the abstention gate converts
into more deferral on the harder distribution.}
\label{fig:ood}
\end{figure}

\subsection{Clinical Risk Category Analysis}
\label{subsec:category-analysis}

\begin{figure}[h]
\centering
\includegraphics[width=0.85\textwidth]{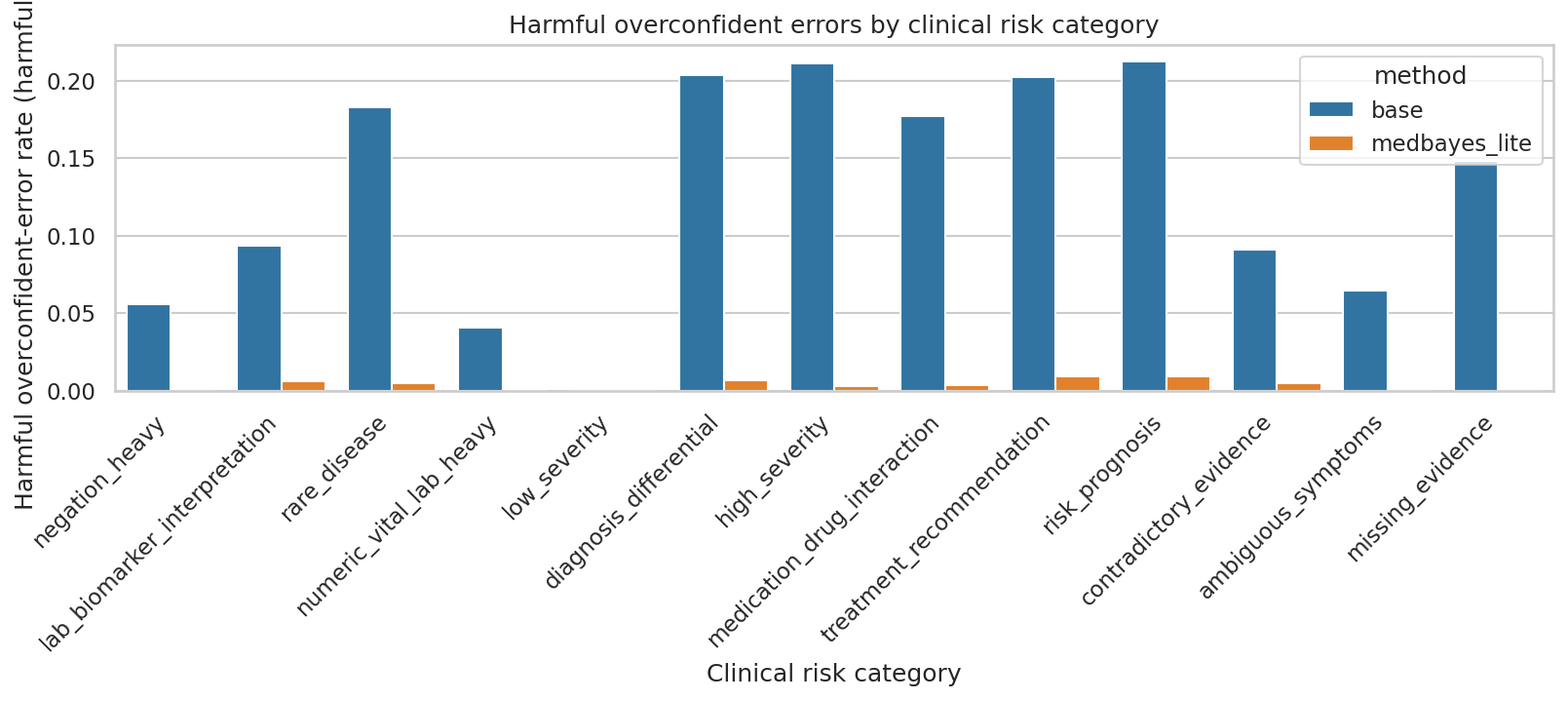}
\caption{Harmful overconfident error rate by rule-based clinical risk category (MedMCQA). Categories are assigned using transparent keyword and pattern rules and are used only as evaluation strata, not as ground-truth clinical labels. Base-model harmful errors concentrate in high-severity, diagnosis, treatment, and medication categories; \ModelName reduces them toward zero.}
\label{fig:risk-categories}
\end{figure}

We assign every item to one or more of 13 rule-based clinical risk categories (Table~\ref{tab:risk_categories}) with severity
weights (medication and drug interaction, diagnosis, treatment, risk and prognosis, laboratory
interpretation, rare disease, ambiguous symptoms, contradictory and missing evidence,
negation-heavy and numeric-heavy text, and high- versus low-severity). Figure~\ref{fig:risk-categories}
shows that base-model harmful overconfidence concentrates in the high-severity, diagnosis,
treatment, and medication categories, where confident error is most costly,
and that \ModelName drives the harmful-error rate in these categories toward zero. This
severity-resolved view is the governance result a clinical reader cares about.

\subsection{Empirical Analysis of CUS}
\label{subsec:cusval}

As a supplementary analysis, we empirically evaluate the proposed clinical summary metrics against the clinical failure metrics they are
meant to summarize, computed across 739 model/method/category groups (Table~\ref{tab:cusval}). CUS
correlates strongly with the harmful overconfident error rate (Pearson $r=0.88$) and with
severity-weighted error ($r=0.90$), and negatively with abstention recall ($r=-0.81$); its weak
correlation with AURC ($r=0.24$) is expected, since CUS is a calibration metric, not a ranking
metric. ZTI, by contrast, has no monotone relationship with coverage (Spearman $r=-0.05$) and
correlates strongly only with AURC ($r=-0.70$), a standard metric it does not improve upon. We
therefore retain CUS as a clinical-calibration summary and demote ZTI to a secondary descriptor.

\begin{table}[h]
\centering
\caption{Empirical analysis of the secondary clinical summary metrics. Correlations are across 739
model/method/category groups. CUS tracks clinical failure; ZTI is threshold-confounded and
redundant with AURC.}
\label{tab:cusval}
\renewcommand{\arraystretch}{1.2}
\small
\begin{tabular}{llc}
\hline
\textbf{Summary} & \textbf{Target metric} & \textbf{Correlation} \\
\hline
CUS & harmful overconfident error rate & Pearson $0.88$ \\
CUS & severity-weighted error & Pearson $0.90$ \\
CUS & abstention recall & Pearson $-0.81$ \\
CUS & AURC & Pearson $0.24$ (weak, expected) \\
\hline
ZTI & AURC & Pearson $-0.70$ \\
ZTI & selective accuracy & Pearson $0.34$ \\
ZTI & coverage & Spearman $-0.05$ (none) \\
\hline
\end{tabular}
\end{table}

\paragraph{Incremental value over ECE.} Because CUS is a severity-weighted calibration error, we further ask whether it adds information beyond standard ECE. We regress each clinical failure metric on ECE alone and on ECE$+$CUS across model/method/category groups (Table~\ref{tab:cus_incremental}). Adding CUS raises the explained variance in the harmful overconfident error rate from $R^2=0.60$ to $0.79$ ($\Delta R^2=0.19$, bootstrap 95\% CI $[0.14, 0.25]$, partial-$F$ $p<10^{-4}$), and in severity-weighted error from $0.73$ to $0.82$ ($\Delta R^2=0.08$, CI $[0.06, 0.11]$). CUS therefore explains additional variance in clinical failure beyond ECE: it captures the severity structure of miscalibration that ECE, which weights all errors equally, does not, while remaining interpretable as a severity-weighted extension of ECE.

\begin{table}[h]
\centering
\caption{Incremental value of CUS over ECE. Nested regressions of each clinical failure
metric on ECE alone versus ECE$+$CUS, across model/method/clinical-risk-category groups.
$\Delta R^2$ is the variance in the failure metric explained by CUS beyond ECE; the interval
is a 2000-sample bootstrap CI, and the $p$-value is a partial-$F$ test for adding CUS. CUS
explains additional variance in clinical failure beyond ECE.}
\label{tab:cus_incremental}
\renewcommand{\arraystretch}{1.15}
\small
\begin{tabular}{lcccccc}
\hline
\textbf{Scope} & \textbf{$n$} & \textbf{$R^2$(ECE)} & \textbf{$R^2$(ECE+CUS)} & \textbf{$\Delta R^2$} & \textbf{$\Delta R^2$ 95\% CI} & \textbf{partial-$F$ $p$} \\
\hline
\multicolumn{7}{l}{\textit{Target: harmful overconfident error rate}} \\
pooled (all groups) & 686 & 0.601 & 0.793 & 0.192 & [0.136, 0.248] & $<$1e-4 \\
model=bert\_base & 78 & 0.321 & 0.461 & 0.140 & [0.037, 0.267] & $<$1e-4 \\
model=biobert & 78 & 0.192 & 0.312 & 0.120 & [0.001, 0.273] & 0.001 \\
model=biobert\_medmcqa & 130 & 0.657 & 0.848 & 0.191 & [0.098, 0.293] & $<$1e-4 \\
model=clinicalbert & 78 & 0.260 & 0.511 & 0.251 & [0.010, 0.451] & $<$1e-4 \\
model=pubmedbert & 78 & 0.391 & 0.477 & 0.086 & [0.019, 0.172] & 0.001 \\
model=pubmedbert\_medmcqa & 130 & 0.529 & 0.812 & 0.284 & [0.170, 0.403] & $<$1e-4 \\
model=qwen2\_5\_7b & 114 & 0.223 & 0.762 & 0.539 & [0.328, 0.693] & $<$1e-4 \\
\hline
\multicolumn{7}{l}{\textit{Target: severity-weighted error}} \\
pooled (all groups) & 686 & 0.734 & 0.818 & 0.084 & [0.057, 0.114] & $<$1e-4 \\
model=bert\_base & 78 & 0.326 & 0.466 & 0.140 & [0.035, 0.269] & $<$1e-4 \\
model=biobert & 78 & 0.285 & 0.333 & 0.048 & [0.001, 0.130] & 0.023 \\
model=biobert\_medmcqa & 130 & 0.795 & 0.878 & 0.082 & [0.036, 0.134] & $<$1e-4 \\
model=clinicalbert & 78 & 0.347 & 0.521 & 0.174 & [0.009, 0.318] & $<$1e-4 \\
model=pubmedbert & 78 & 0.402 & 0.489 & 0.087 & [0.018, 0.173] & 0.001 \\
model=pubmedbert\_medmcqa & 130 & 0.670 & 0.820 & 0.150 & [0.096, 0.209] & $<$1e-4 \\
model=qwen2\_5\_7b & 114 & 0.551 & 0.800 & 0.249 & [0.096, 0.403] & $<$1e-4 \\
\hline
\end{tabular}
\end{table}

\subsection{Generative Contrast and Negative Findings}
\label{subsec:negative}

\paragraph{Risk ranking does not improve (AURC).}
The improvements above are to the confidence \emph{scale}, not to the \emph{ordering} of
predictions by reliability. In Table~\ref{tab:main_results} the AURC of \ModelName is
statistically indistinguishable from the base model in three of four settings (paired bootstrap
$p$ between $0.6$ and $0.96$) and marginally worse in one, and the error-detection AUROC is no
higher. Consistent with this, on PubMedQA \emph{deep ensembles} achieve the best AURC ($0.345$
to $0.363$ across the encoder family) while \ModelName does not (Table~\ref{tab:pubmedqa}).
Ranking gains come from ensembling; calibration gains come from temperature scaling or Monte
Carlo averaging; \ModelName combines the latter two and should not be claimed to improve ranking.

\paragraph{Component ablation.} Table~\ref{tab:ablation_components} isolates the components of the layer on MedMCQA (PubMedBERT). Temperature calibration is the dominant lever (ECE $0.255\to0.081$); MC-dropout averaging contributes a smaller, complementary improvement ($0.255\to0.092$); the abstention-only variant is numerically identical to the base predictor, since abstention is a threshold policy applied to the underlying confidence; and the full layer (MC-dropout, temperature, and abstention) is best (ECE $0.030$, harmful $0.002$). AURC is unchanged across all variants. The uncertainty-weighted attention component is inactive in these runs ($\lambda=0$) and is not independently isolated (Section~\ref{sec:limitations}); the gains are attributable to calibration and abstention.

\begin{table}[h]
\centering
\caption{Component ablation on MedMCQA (PubMedBERT). Temperature calibration is the dominant
lever for calibration and harmful-error reduction; MC-dropout averaging adds a smaller
improvement; the abstention-only variant equals the base predictor because abstention is a
threshold policy applied to the (here uncalibrated) confidence. The full layer is MC-dropout
plus temperature plus abstention. AURC is essentially unchanged across variants.}
\label{tab:ablation_components}
\renewcommand{\arraystretch}{1.15}
\small
\begin{tabular}{lcccccc}
\hline
\textbf{Variant} & \textbf{Acc} & \textbf{ECE}$\downarrow$ & \textbf{harmful@0.8}$\downarrow$ & \textbf{SelAcc@0.7}$\uparrow$ & \textbf{Cov@0.7} & \textbf{AURC}$\downarrow$ \\
\hline
Base & 0.465 & 0.255 & 0.056 & 0.571 & 0.541 & 0.376 \\
MC dropout only & 0.455 & 0.092 & 0.010 & 0.734 & 0.198 & 0.395 \\
Temperature scaling only & 0.465 & 0.081 & 0.012 & 0.721 & 0.217 & 0.375 \\
Abstention only & 0.465 & 0.255 & 0.056 & 0.571 & 0.541 & 0.376 \\
Full (MC + temperature + abstention) & 0.455 & 0.030 & 0.002 & 0.811 & 0.102 & 0.395 \\
\hline
\end{tabular}
\end{table}

\paragraph{No benefit when the model is already calibrated.}
Table~\ref{tab:pubmedqa} also shows a null result. Our locally fine-tuned PubMedQA encoders are
already well calibrated (ECE $0.06$ to $0.08$) with no measurable harmful overconfidence, and
\ModelName leaves these properties essentially unchanged. The framework helps only when there is
harmful overconfidence to remove.

\begin{table}[h]
\centering
\caption{Null-result control on PubMedQA, where the fine-tuned encoders are already well
calibrated. \ModelName neither helps nor hurts calibration; deep ensembles give the best AURC.
Harmful overconfident error rate is $0.000$ for all rows.}
\label{tab:pubmedqa}
\renewcommand{\arraystretch}{1.2}
\small
\begin{tabular}{lccc}
\hline
\textbf{Model} & \textbf{base ECE / AURC} & \textbf{MedBayes-Lite ECE / AURC} & \textbf{ensemble AURC} \\
\hline
BERT-base & 0.069 / 0.376 & 0.074 / 0.385 & \textbf{0.351} \\
BioBERT & 0.069 / 0.376 & 0.074 / 0.399 & \textbf{0.345} \\
Bio\_ClinicalBERT & 0.060 / 0.394 & 0.078 / 0.428 & \textbf{0.346} \\
PubMedBERT & 0.075 / 0.371 & 0.071 / 0.382 & \textbf{0.363} \\
\hline
\end{tabular}
\end{table}

\paragraph{Generative models are more dangerous, and self-consistency does not fix it.}
Table~\ref{tab:generative} evaluates Qwen2.5-7B-Instruct in the sampling-only mode. The
generative model is far more overconfident than the encoders: on PubMedQA, self-consistency
gives an ECE of $0.445$ and a harmful overconfident error rate of $0.360$, and it does not
improve over semantic entropy because the samples agree on the wrong answer. Eliciting a
\emph{verbalized} confidence calibrates substantially better (PubMedQA ECE $0.167$, harmful
$0.110$). Notably, on PubMedQA the model's uncertainty ranks its errors well (AUROC of error
detection $0.78$), so the failure is a miscalibrated confidence scale, precisely the regime a
calibration-and-abstention governance layer is meant to address. This contrast motivates
governance rather than demonstrating a uniform improvement.

\begin{table}[h]
\centering
\caption{Generative model (Qwen2.5-7B-Instruct), sampling-only uncertainty. The model is
strongly overconfident; self-consistency does not repair it, and verbalized confidence
calibrates better.}
\label{tab:generative}
\renewcommand{\arraystretch}{1.2}
\small
\begin{tabular}{llccc}
\hline
\textbf{Dataset} & \textbf{Method} & \textbf{Accuracy} & \textbf{ECE}$\downarrow$ & \textbf{harmful@0.8}$\downarrow$ \\
\hline
PubMedQA & self-consistency & 0.525 & 0.445 & 0.360 \\
PubMedQA & semantic entropy & 0.525 & 0.417 & 0.345 \\
PubMedQA & verbalized & 0.650 & \textbf{0.167} & \textbf{0.110} \\
MedQA-USMLE & self-consistency & 0.685 & 0.307 & 0.265 \\
MedQA-USMLE & verbalized & 0.690 & \textbf{0.167} & 0.260 \\
MedMCQA & self-consistency & 0.495 & 0.484 & 0.115 \\
MedMCQA & verbalized & 0.480 & \textbf{0.355} & 0.125 \\
\hline
\end{tabular}
\end{table}

\paragraph{Sensitivity to clinical perturbations is partial.}
Table~\ref{tab:perturb} reports how \ModelName responds to controlled corruptions of MedMCQA
items. It raises uncertainty and abstention appropriately for irrelevant distractors (mean
confidence drop $0.051$) and contradictory findings ($0.033$), but it is essentially insensitive
to corruption of numeric or vital-sign values (confidence change $-0.001$) and to vague phrasing
($0.000$). The layer governs aggregate predictive uncertainty and does not detect local numeric
implausibility, a limitation that matters for dosing and laboratory interpretation.

\begin{table}[h]
\centering
\caption{Response of \ModelName to controlled clinical perturbations (MedMCQA, $n=300$).
Positive confidence drop and abstention increase indicate appropriate added caution.}
\label{tab:perturb}
\renewcommand{\arraystretch}{1.2}
\small
\begin{tabular}{lccc}
\hline
\textbf{Perturbation} & \textbf{Confidence drop} & \textbf{Abstention increase} & \textbf{Flip rate} \\
\hline
add irrelevant distractor & 0.051 & 0.073 & 0.37 \\
add contradictory symptom & 0.033 & 0.060 & 0.34 \\
remove key evidence & 0.015 & 0.037 & 0.19 \\
remove lab value & 0.006 & 0.017 & 0.16 \\
alter numeric value & $-0.001$ & 0.013 & 0.14 \\
vague evidence & 0.000 & 0.000 & 0.16 \\
\hline
\end{tabular}
\end{table}

\subsection{Computational Efficiency}
\label{subsec:efficiency}

\ModelName adds no trainable parameters and reuses the backbone across stochastic passes, so its
parameter count and memory footprint match the base model; the primary cost is the $M$ Monte
Carlo forward passes, so latency grows approximately linearly with $M$.
Table~\ref{tab:efficiency_benchmark} reports measured latency and memory on a Bio\_ClinicalBERT
backbone. Memory is constant at $0.43$ GB across Monte Carlo settings, in contrast to a
five-member deep ensemble ($541$M parameters, $2.85$ GB). We note that, unlike memory, compute
is not negligible: latency roughly doubles at $M{=}10$ relative to a single pass, so the
``Lite'' property refers to parameter and memory efficiency, not a flat compute overhead. The
exact figures are hardware-dependent.

\begin{table}[h]
\centering
\caption{Computational efficiency on a Bio\_ClinicalBERT backbone. ``MC/Ens.'' is the number of
Monte Carlo samples (\ModelName, SWAG) or ensemble members. \ModelName keeps the backbone
parameter count and a constant memory footprint; latency scales approximately linearly with the
number of Monte Carlo samples.}
\label{tab:efficiency_benchmark}
\renewcommand{\arraystretch}{1.2}
\begin{tabular}{lcccc}
\toprule
\textbf{Model Variant} & \textbf{Params} & \textbf{MC/Ens.} & \textbf{Latency (ms)} & \textbf{GPU Mem (GB)} \\
\midrule
Bio\_ClinicalBERT (Baseline) & 108M & --  & 32.04 & 0.43 \\
\textit{MedBayes-Lite} (MC=5)   & 108M & 5  & 32.61 & 0.43 \\
\textit{MedBayes-Lite} (MC=10)  & 108M & 10 & 60.13 & 0.43 \\
\textit{MedBayes-Lite} (MC=20)  & 108M & 20 & 119.08 & 0.43 \\
\textit{MedBayes-Lite} (MC=50)  & 108M & 50 & 298.54 & 0.43 \\
SWAG (n=10 samples)          & 108M & 10 & 103.93 & 0.98 \\
Deep Ensemble (N=5)          & 541M & -- & 31.15 & 2.85 \\
\bottomrule
\end{tabular}
\end{table}
\section{Discussion}
\label{sec:discussion}

The primary contribution of \textit{MedBayes-Lite} is not a new uncertainty estimator, but a practical governance mechanism that converts uncertainty estimates into calibrated abstention behavior. Read this way, \textit{MedBayes-Lite} is best understood as a governance layer, not as an uncertainty estimator that dominates prior work. Read this way, the evidence is consistent and
interpretable.

\subsection{What the layer does}

First, it reduces confident high-severity errors on the benchmark clinical QA tasks. On clinical multiple-choice QA it
reduces calibration error substantially (0.23 to 0.33 absolute ECE) and removes nearly all
harmful overconfident errors, those that are simultaneously confident, incorrect, and clinically
high-severity. Second, it supports appropriate abstention: at a fixed acceptance threshold the
accuracy of accepted predictions rises sharply (for example from 0.57 to 0.82 on MedMCQA with
PubMedBERT), and abstention recall, the fraction of incorrect predictions that are deferred,
reaches 0.97 to 0.99. Third, and most relevant to settings with distribution shift, the benefit is largest under
domain shift. When models trained on MedMCQA are evaluated on MedQA-USMLE, the uncalibrated
baseline becomes acutely dangerous, committing confident errors on roughly 21\% of high-severity
items; the governance layer reduces this to near zero and roughly halves the calibration drift.
A clinical model that is most overconfident exactly when it is most likely to be wrong is the
central failure this layer is designed to contain.

\subsection{What the layer does not do}

We are explicit about the boundaries of these claims. The layer does not improve the model's
ability to \emph{rank} its predictions by reliability: the area under the risk-coverage curve is
statistically unchanged in three of four multiple-choice settings and marginally worse in the
fourth, and the area under the error-detection ROC curve is no higher. The improvement is a
change to the confidence \emph{scale}, not to the \emph{ordering} of predictions, and a perfectly
calibrated and a poorly calibrated model can share the same risk-coverage curve. Consequently,
temperature scaling, a far cheaper post-hoc method, matches most of the calibration gain, and
deep ensembles achieve better risk ranking than our layer. The safety we report is also bought
at a measurable cost in coverage: the model answers less often. And on models that are already
well calibrated, such as our fine-tuned PubMedQA encoders, the layer provides no benefit because
there is no harmful overconfidence left to remove.

A contrast with an open-weight generative model sharpens the motivation rather than the method.
Qwen2.5-7B-Instruct is markedly more overconfident than the encoders, confidently wrong on about
36\% of high-severity PubMedQA items, and self-consistency does not repair this because the
samples agree on the wrong answer. Notably, the model's uncertainty does separate its errors well
in a ranking sense, so the failure is purely a miscalibrated confidence scale, exactly the regime
a calibration-and-abstention governance layer is meant to address. Eliciting a verbalized
confidence calibrates better than self-consistency, which is itself a useful, if narrow, finding
for sampling-only deployments.

The empirical evidence suggests that \textit{MedBayes-Lite} should not be viewed as a replacement for strong calibration baselines or ensemble uncertainty methods. Temperature scaling achieves comparable calibration at lower cost, while deep ensembles provide stronger risk ranking. The primary contribution of \textit{MedBayes-Lite} is instead a practical uncertainty-governance mechanism that combines calibration and confidence-guided abstention to reduce harmful overconfident clinical errors.

\subsection{Clinical relevance}

The practical value of the layer is in the deployment interface it creates. By deferring
predictions whose calibrated confidence falls below a threshold, it converts a
silent answer engine into a system that escalates uncertain, high-stakes cases for human review.
This is most useful under the conditions clinicians actually face at deployment time: new
institutions, new note styles, and case mixes that differ from training. It does not replace
clinical judgment, and it should not be presented as a guarantee of safety; it reduces the rate
at which a model commits confident high-severity errors, and it does so without retraining or
added parameters, which matters when the underlying predictor cannot be modified.

\subsection{Relation to standard metrics}

Our metric analysis cautions against summarizing clinical uncertainty with a single composite
score. The Clinical Uncertainty Score is useful: it is a severity-weighted calibration metric and
it tracks harmful overconfidence closely (Pearson $r\approx0.88$). Its limitation is that it is, by construction, a severity-weighted extension of ECE rather than an independent metric, and its severity weights are a modeling choice; we therefore report it only as a supplementary summary and validate its robustness to those weights in the Supplementary Material. The Zero-shot Trustworthiness
Index, by contrast, is threshold-confounded and adds little beyond the standard area under the
risk-coverage curve, so we report it only as a secondary descriptor. More generally, the right
evaluation for a governance layer pairs calibration metrics (ECE, CUS) and selective-prediction
metrics (AURC, selective accuracy, coverage, abstention recall) rather than collapsing them,
because the layer improves the former without improving the latter.

\section{Limitations}
\label{sec:limitations}

We report the following limitations so that the scope of our claims is unambiguous.

\paragraph{No improvement in risk ranking (AURC).} The layer changes the confidence scale but not
the ordering of predictions by reliability. The area under the risk-coverage curve is
statistically unchanged in three of four multiple-choice settings ($p$ between 0.6 and 0.96) and
marginally worse in the fourth, and the error-detection AUROC is no higher. Temperature scaling
matches the calibration gain at lower cost, and deep ensembles achieve better risk ranking on
PubMedQA. Claims for this layer should be confined to calibration, harmful-error reduction, and
abstention behavior, not selective-prediction superiority.

\paragraph{Null result on already-calibrated models.} On our fine-tuned PubMedQA encoders, which
are already well calibrated (ECE around 0.06 to 0.08 with no measurable harmful overconfidence),
the layer provides no benefit. The framework helps only when there is harmful overconfidence to
remove; it is not a universal improvement.

\paragraph{Coverage cost.} The safety gains are bought by answering less often. At a fixed
acceptance threshold, coverage can fall substantially (for example from 0.54 to 0.10 on MedMCQA
with PubMedBERT). For a hard exam-style task this conservative behavior is defensible, but in
deployment the abstention threshold must be tuned to an acceptable workload, and the
coverage-reliability trade-off must be reported, not hidden.

\paragraph{Insensitivity to local numeric corruption.} In controlled perturbation tests the layer
raised uncertainty appropriately for irrelevant distractors and contradictory findings, but it
did not respond to clinically implausible numeric or vital-sign values (mean confidence change
near zero) or to vague phrasing. The layer governs aggregate predictive uncertainty and does not
detect local numeric implausibility, which is precisely the failure mode that matters for dosing
and laboratory interpretation.

\paragraph{The optional attention extension is not evaluated.} Empirically the layer behaves as Monte Carlo dropout plus calibration plus a confidence-guided abstention gate. The optional uncertainty-weighted attention extension (Appendix~\ref{app:attention}) was inactive ($\lambda=0$) in all reported experiments, so we draw no empirical conclusions about it; evaluating alternative uncertainty estimation strategies remains future work.

\paragraph{Datasets and access constraints.} The evaluation covers MedMCQA, MedQA-USMLE, and
PubMedQA. The domain-shift evidence rests on a single source-to-target transfer (MedMCQA to MedQA-USMLE), so the magnitude of the out-of-distribution benefit should be read as indicative rather than exhaustive. We did not evaluate on MIMIC-III electronic health records, which require credentialed
access and were outside the scope of this study; the in-hospital mortality task is left as a
clearly separable future axis, and the released code includes a credentialed-access adapter for
it. The generative evaluation uses a single open-weight model on a sampled subset rather than a
broad sweep, and we did not evaluate proprietary API models.

\paragraph{Experimental scale.} Results use three random seeds rather than five; for the
pretrained multiple-choice checkpoints the deterministic baselines have no seed variance, and
confidence intervals for all methods are obtained by per-sample bootstrap. The generative
evaluation uses 200 items per dataset for cost reasons. We expect the qualitative conclusions to
be stable, but the exact magnitudes may shift with larger runs.

\paragraph{Clinical deployment is not evaluated.} All results are retrospective, on benchmark
data. We make no claim about prospective clinical safety, workflow integration, or effect on
clinician behavior. Demonstrating that confident high-severity errors are reduced on benchmarks is
a necessary precondition for, but not evidence of, safe clinical use.

\section{Conclusion}
\label{sec:conclusion}

We presented \ModelName, a clinical uncertainty governance layer for pretrained clinical
predictors. Rather than adding parameters or retraining, it converts inference-time Monte Carlo
dropout uncertainty into a calibrated, confidence-guided abstention policy that defers
low-confidence predictions for human review. Across clinical multiple-choice question answering,
the layer substantially improved calibration and drove harmful overconfident clinical errors
toward zero, with the largest benefit under medical domain shift, where it contained the confident
high-severity errors that an uncalibrated baseline commits. We introduced and empirically evaluated the
Clinical Uncertainty Score as a severity-weighted calibration summary of harmful overconfidence,
and we showed that the previously proposed ZTI is threshold-confounded and should be demoted.

We also retain the negative findings that bound these claims: the layer does not improve risk
ranking (its risk-coverage area is statistically unchanged, temperature scaling matches its
calibration, and deep ensembles rank risk better), it provides no benefit on already
well-calibrated models, it carries a coverage cost, it does not detect local numeric
implausibility, and the optional attention pathway was inactive in all reported experiments. A
contrast with an open-weight generative model showed that naive self-consistency does not confer
safety, reinforcing the case for explicit governance.

These results reframe lightweight Bayesian inference for clinical models as a governance problem:
the goal is not a better uncertainty score in the abstract, but a practical mechanism that
converts uncertainty into calibrated, risk-aware deferral. Future work includes broader domain-shift evaluation across additional source-target pairs, additional healthcare benchmarks, alternative uncertainty estimators, prospective clinical studies, and evaluation on credentialed
electronic-health-record tasks such as MIMIC-III in-hospital mortality, detectors for local
numeric and dosing implausibility, and prospective study of how confidence-guided abstention
affects clinician decision-making.

\section*{Author Contributions}

E.H. conceived the study, designed the methodology, conducted the experiments, implemented the framework, analyzed the results, and wrote the main manuscript. M.M.H.N. contributed to the development of MedBayes-Lite and assisted with the experiments. M.S. contributed to data analysis, visualization, and manuscript editing. T.N. and B.W.S. contributed to manuscript editing, revision, and critical feedback. R.R., S.N., and N.Y. supervised the study and reviewed the manuscript. All authors reviewed and approved the final manuscript.

\section*{Acknowledgement}

We would like to express our special thanks to Dr. Yongchao Huang, Assistant Professor at the University of Aberdeen, UK, for reviewing the methodology of this paper and providing valuable feedback. His insightful comments and suggestions greatly helped improve the quality and rigor of this work.

\section*{Ethics Statement}
This study is based entirely on publicly available, de-identified biomedical QA datasets (MedMCQA, MedQA-USMLE, PubMedQA) and does not involve direct interaction with human participants. Therefore, formal ethics approval was not required. These benchmarks do not contain identifiable patient information. MIMIC-III electronic health records were not used in this study. All data handling complied with applicable data privacy guidelines and ethical standards for secondary data analysis.

\section*{Declaration of Competing Interest}
The authors declare that they have no known competing financial interests or personal relationships that could have appeared to influence the work reported in this paper.

\section*{Data Availability}

All datasets used in this study are de-identified. PubMedQA \cite{jin2019pubmedqa} and MedQA \cite{yang2025llmmedqaenhancingmedicalquestion} are publicly downloadable from their official repositories as referenced in the manuscript. MIMIC-III \cite{johnson2016mimic} is accessible via PhysioNet under credentialed access but was not evaluated in this study; the released code includes a credentialed-access adapter for the in-hospital mortality task as a future axis. Processed evaluation subsets, per-sample predictions, metric tables, and plots generated during this study are released with the accompanying code.

\section*{Funding Statement}
This research did not receive any specific grant from funding agencies in the public, commercial, or not-for-profit sectors.

\clearpage
\bibliographystyle{IEEEtran}
\bibliography{medbayes/references}

@article{savage2024large,
  author  = {Savage, Thomas and Wang, John and Gallo, Robert and Boukil, Abdessalem and Patel, Vishwesh and Ahmad Safavi-Naini, Seyed Amir and Soroush, Ali and Chen, Jonathan H.},
  title   = {Large language model uncertainty measurement and calibration for medical diagnosis and treatment},
  journal = {medRxiv},
  pages   = {2024--06},
  year    = {2024},
  publisher = {Cold Spring Harbor Laboratory Press}
}

@article{salvi2025explainability,
  author  = {Salvi, Massimo and Seoni, Silvia and Campagner, Andrea and Gertych, Arkadiusz and Acharya, U. Rajendra and Molinari, Filippo and Cabitza, Federico},
  title   = {Explainability and uncertainty: Two sides of the same coin for enhancing the interpretability of deep learning models in healthcare},
  journal = {International Journal of Medical Informatics},
  pages   = {105846},
  year    = {2025},
  publisher = {Elsevier}
}

@article{catak2024uncertainty,
  author  = {Catak, Ferhat Ozgur and Kuzlu, Murat},
  title   = {Uncertainty quantification in large language models through convex hull analysis},
  journal = {Discover Artificial Intelligence},
  volume  = {4},
  number  = {1},
  pages   = {1--14},
  year    = {2024},
  publisher = {Springer}
}

@inproceedings{10.5555/3737916.3738407,
author = {Ye, Fanghua and Yang, Mingming and Pang, Jianhui and Wang, Longyue and Wong, Derek F. and Yilmaz, Emine and Shi, Shuming and Tu, Zhaopeng},
title = {Benchmarking LLMs via uncertainty quantification},
year = {2024},
isbn = {9798331314385},
publisher = {Curran Associates Inc.},
address = {Red Hook, NY, USA},
booktitle = {Proceedings of the 38th International Conference on Neural Information Processing Systems},
articleno = {491},
numpages = {30},
location = {Vancouver, BC, Canada},
series = {NIPS '24}
}

@article{lievin2024can,
  author  = {Li{\'e}vin, Valentin and Hother, Christoffer Egeberg and Motzfeldt, Andreas Geert and Winther, Ole},
  title   = {Can large language models reason about medical questions?},
  journal = {Patterns},
  volume  = {5},
  number  = {3},
  year    = {2024},
  publisher = {Elsevier}
}

@inproceedings{joo2020being,
  title={Being bayesian about categorical probability},
  author={Joo, Taejong and Chung, Uijung and Seo, Min-Gwan},
  booktitle={International conference on machine learning},
  pages={4950--4961},
  year={2020},
  organization={PMLR}
}

@inproceedings{lakshminarayanan2017simple,
  author    = {Lakshminarayanan, B. and Pritzel, A. and Blundell, C.},
  title     = {Simple and Scalable Predictive Uncertainty Estimation using Deep Ensembles},
  booktitle = {Advances in Neural Information Processing Systems},
  volume    = {30},
  year      = {2017}
}

@inproceedings{gal2016dropout,
  author    = {Gal, Yarin and Ghahramani, Zoubin},
  title     = {Dropout as a Bayesian Approximation: Representing Model Uncertainty in Deep Learning},
  booktitle = {International Conference on Machine Learning},
  pages     = {1050--1059},
  year      = {2016},
  organization = {PMLR}
}

@inproceedings{10.5555/3305381.3305518,
author = {Guo, Chuan and Pleiss, Geoff and Sun, Yu and Weinberger, Kilian Q.},
title = {On calibration of modern neural networks},
year = {2017},
publisher = {JMLR.org},
booktitle = {Proceedings of the 34th International Conference on Machine Learning - Volume 70},
pages = {1321–1330},
numpages = {10},
location = {Sydney, NSW, Australia},
series = {ICML'17}
}

@article{thirunavukarasu2023large,
  author  = {Thirunavukarasu, Arun James and Ting, Darren Shu Jeng and Elangovan, Kabilan and Gutierrez, Laura and Tan, Ting Fang and Ting, Daniel Shu Wei},
  title   = {Large Language Models in Medicine},
  journal = {Nature Medicine},
  volume  = {29},
  number  = {8},
  pages   = {1930--1940},
  year    = {2023},
  publisher = {Nature Publishing Group US New York}
}

@article{alberts2023large,
  author  = {Alberts, Ian L. and Mercolli, Lorenzo and Pyka, Thomas and Prenosil, George and Shi, Kuangyu and Rominger, Axel and Afshar-Oromieh, Ali},
  title   = {Large language models ({LLM}) and ChatGPT: what will the impact on nuclear medicine be?},
  journal = {European Journal of Nuclear Medicine and Molecular Imaging},
  volume  = {50},
  number  = {6},
  pages   = {1549--1552},
  year    = {2023},
  publisher = {Springer}
}

@article{lu2024large,
  author  = {Lu, Zhiyong and Peng, Yifan and Cohen, Trevor and Ghassemi, Marzyeh and Weng, Chunhua and Tian, Shubo},
  title   = {Large Language Models in Biomedicine and Health: Current Research Landscape and Future Directions},
  journal = {Journal of the American Medical Informatics Association},
  volume  = {31},
  number  = {9},
  pages   = {1801--1811},
  year    = {2024},
  publisher = {Oxford University Press}
}

@article{mora2024trustworthy,
  author  = {Mora-Cantallops, Mar{\c{c}}al and Garc{\'\i}a-Barriocanal, Elena and Sicilia, Miguel-{\'A}ngel},
  title   = {Trustworthy {AI} Guidelines in Biomedical Decision-Making Applications: A Scoping Review},
  journal = {Big Data and Cognitive Computing},
  volume  = {8},
  number  = {7},
  pages   = {73},
  year    = {2024},
  publisher = {MDPI}
}

@article{abdar2021review,
  author  = {Abdar, Moloud and Pourpanah, Farhad and Hussain, Sadiq and Rezazadegan, Dana and Liu, Li and Ghavamzadeh, Mohammad and Fieguth, Paul and Cao, Xiaochun and Khosravi, Abbas and Acharya, U. Rajendra and others},
  title   = {A Review of Uncertainty Quantification in Deep Learning: Techniques, Applications and Challenges},
  journal = {Information Fusion},
  volume  = {76},
  pages   = {243--297},
  year    = {2021},
  publisher = {Elsevier}
}

@article{hullermeier2021aleatoric,
  author  = {H{\"u}llermeier, Eyke and Waegeman, Willem},
  title   = {Aleatoric and Epistemic Uncertainty in Machine Learning: An Introduction to Concepts and Methods},
  journal = {Machine Learning},
  volume  = {110},
  number  = {3},
  pages   = {457--506},
  year    = {2021},
  publisher = {Springer}
}

@misc{yang2025llmmedqaenhancingmedicalquestion,
  author       = {Yang, Hang and Chen, Hao and Guo, Hui and Chen, Yineng and Lin, Ching-Sheng and Hu, Shu and Hu, Jinrong and Wu, Xi and Wang, Xin},
  title        = {{LLM}-MedQA: Enhancing Medical Question Answering Through Case Studies in Large Language Models},
  year         = {2025},
  eprint       = {2501.05464},
  archivePrefix= {arXiv},
  primaryClass = {cs.CL},
  url          = {https://arxiv.org/abs/2501.05464}
}

@inproceedings{jin2019pubmedqa,
  title={Pubmedqa: A dataset for biomedical research question answering},
  author={Jin, Qiao and Dhingra, Bhuwan and Liu, Zhengping and Cohen, William and Lu, Xinghua},
  booktitle={Proceedings of the 2019 conference on empirical methods in natural language processing and the 9th international joint conference on natural language processing (EMNLP-IJCNLP)},
  pages={2567--2577},
  year={2019}
}

@misc{hendrycks2018baselinedetectingmisclassifiedoutofdistribution,
  title         = {A Baseline for Detecting Misclassified and Out-of-Distribution Examples in Neural Networks},
  author        = {Hendrycks, Dan and Gimpel, Kevin},
  year          = {2018},
  eprint        = {1610.02136},
  archivePrefix = {arXiv},
  primaryClass  = {cs.NE},
  url           = {https://arxiv.org/abs/1610.02136}
}

@inproceedings{niculescu2005predicting,
  title     = {Predicting good probabilities with supervised learning},
  author    = {Niculescu-Mizil, Alexandru and Caruana, Rich},
  booktitle = {Proceedings of the 22nd International Conference on Machine Learning},
  pages     = {625--632},
  year      = {2005}
}

@article{maddox2019simple,
  title={A simple baseline for bayesian uncertainty in deep learning},
  author={Maddox, Wesley J and Izmailov, Pavel and Garipov, Timur and Vetrov, Dmitry P and Wilson, Andrew Gordon},
  journal={Advances in neural information processing systems},
  volume={32},
  year={2019}
}

@article{johnson2016mimic,
  title={MIMIC-III, a freely accessible critical care database},
  author={Johnson, Alistair EW and Pollard, Tom J and Shen, Lu and Lehman, Li-wei H and Feng, Mengling and Ghassemi, Mohammad and Moody, Benjamin and Szolovits, Peter and Anthony Celi, Leo and Mark, Roger G},
  journal={Scientific data},
  volume={3},
  number={1},
  pages={1--9},
  year={2016},
  publisher={Nature Publishing Group}
}

@article{xia2024uncertaintyaware,
  title   = {Uncertainty-Aware Health Diagnostics via Class-Balanced Evidential Deep Learning},
  author  = {Xia, T. and Dang, T. and Han, J. and Qendro, L. and Mascolo, C.},
  journal = {IEEE Journal of Biomedical and Health Informatics},
  volume  = {28},
  number  = {11},
  pages   = {6417--6428},
  year    = {2024},
  doi     = {10.1109/JBHI.2024.3360002},
  url     = {https://doi.org/10.1109/JBHI.2024.3360002}
}

@article{neupane2025medinsight,
  title={Medinsight: A multi-source context augmentation framework for generating patient-centric medical responses using large language models},
  author={Neupane, Subash and Mitra, Shaswata and Mittal, Sudip and Gaur, Manas and Golilarz, Noorbakhsh Amiri and Rahimi, Shahram and Amirlatifi, Amin},
  journal={ACM Transactions on Computing for Healthcare},
  volume={6},
  number={2},
  pages={1--19},
  year={2025},
  publisher={ACM New York, NY}
}

@inproceedings{neupane2024clinicsum,
  title={CLINICSUM: Utilizing language models for generating clinical summaries from patient-doctor conversations},
  author={Neupane, Subash and Tripathi, Himanshu and Mitra, Shaswata and Bozorgzad, Sean and Mittal, Sudip and Rahimi, Shahram and Amirlatifi, Amin},
  booktitle={2024 IEEE International Conference on Big Data (BigData)},
  pages={5050--5059},
  year={2024},
  organization={IEEE}
}

@article{ong2025large,
  title={Large language model as clinical decision support system augments medication safety in 16 clinical specialties},
  author={Ong, Jasmine Chiat Ling and Jin, Liyuan and Elangovan, Kabilan and San Lim, Gilbert Yong and Lim, Daniel Yan Zheng and Sng, Gerald Gui Ren and Ke, Yu He and Tung, Joshua Yi Min and Zhong, Ryan Jian and Koh, Christopher Ming Yao and others},
  journal={Cell Reports Medicine},
  volume={6},
  number={10},
  year={2025},
  publisher={Elsevier}
}

@article{nerella2024transformers,
  title={Transformers and large language models in healthcare: A review},
  author={Nerella, Subhash and Bandyopadhyay, Sabyasachi and Zhang, Jiaqing and Contreras, Miguel and Siegel, Scott and Bumin, Aysegul and Silva, Brandon and Sena, Jessica and Shickel, Benjamin and Bihorac, Azra and others},
  journal={Artificial intelligence in medicine},
  volume={154},
  pages={102900},
  year={2024},
  publisher={Elsevier}
}

@article{arslan2025evaluating,
  title={Evaluating LLM-based generative AI tools in emergency triage: A comparative study of ChatGPT Plus, Copilot Pro, and triage nurses},
  author={Arslan, B and Nuhoglu, C and Satici, MO and Altinbilek, E},
  journal={The American Journal of Emergency Medicine},
  volume={89},
  pages={174--181},
  year={2025},
  publisher={Elsevier}
}

@article{mcduff2025towards,
  title={Towards accurate differential diagnosis with large language models},
  author={McDuff, Daniel and Schaekermann, Mike and Tu, Tao and Palepu, Anil and Wang, Amy and Garrison, Jake and Singhal, Karan and Sharma, Yash and Azizi, Shekoofeh and Kulkarni, Kavita and others},
  journal={Nature},
  pages={1--7},
  year={2025},
  publisher={Nature Publishing Group UK London}
}

@article{oliveira2025development,
  title={Development and evaluation of a clinical note summarization system using large language models},
  author={Oliveira, Juliana Damasio and Santos, Henrique DP and Ulbrich, Ana Helena DPS and Couto, Julia Colleoni and Arocha, Marcelo and Santos, Joaquim and Costa, Manuela Martins and Faccio, Daniela and Tabalipa, Fabio O and Nogueira, Rodrigo F},
  journal={Communications medicine},
  volume={5},
  number={1},
  pages={376},
  year={2025},
  publisher={Nature Publishing Group UK London}
}

@article{sridharan2024unlocking,
  title={Unlocking the potential of advanced large language models in medication review and reconciliation: A proof-of-concept investigation},
  author={Sridharan, Kannan and Sivaramakrishnan, Gowri},
  journal={Exploratory Research in Clinical and Social Pharmacy},
  volume={15},
  pages={100492},
  year={2024},
  publisher={Elsevier}
}

@article{li2025streamlining,
  title={Streamlining evidence based clinical recommendations with large language models},
  author={Li, Dubai and Jiang, Nan and Huang, Kangping and Tu, Ruiqi and Ouyang, Shuyu and Yu, Huayu and Qiao, Lin and Yu, Chen and Zhou, Tianshu and Tong, Danyang and others},
  journal={npj Digital Medicine},
  year={2025},
  publisher={Nature Publishing Group UK London}
}

@inproceedings{vazhentsev2022uncertainty,
  title={Uncertainty estimation of transformer predictions for misclassification detection},
  author={Vazhentsev, Artem and Kuzmin, Gleb and Shelmanov, Artem and Tsvigun, Akim and Tsymbalov, Evgenii and Fedyanin, Kirill and Panov, Maxim and Panchenko, Alexander and Gusev, Gleb and Burtsev, Mikhail and others},
  booktitle={Proceedings of the 60th Annual Meeting of the Association for Computational Linguistics (Volume 1: Long Papers)},
  pages={8237--8252},
  year={2022}
}

@article{sensoy2018evidential,
  title={Evidential deep learning to quantify classification uncertainty},
  author={Sensoy, Murat and Kaplan, Lance and Kandemir, Melih},
  journal={Advances in neural information processing systems},
  volume={31},
  year={2018}
}

@article{gao2025comprehensive,
  title={A comprehensive survey on evidential deep learning and its applications},
  author={Gao, Junyu and Chen, Mengyuan and Xiang, Liangyu and Xu, Changsheng},
  journal={IEEE Transactions on Pattern Analysis and Machine Intelligence},
  year={2025},
  publisher={IEEE}
}


\clearpage
\appendix

\section{Optional Uncertainty-Weighted Attention (Not Evaluated in This Paper)}
\label{app:attention}
\label{sec:uncertainty-adjusted}

When internal transformer states are accessible, token-level uncertainty can optionally modulate attention. From MC-dropout embedding samples $\{h^{(m)}(x)\}$ with empirical mean $\hat{\mu}(x)$ and covariance $\hat{\Sigma}(x)$, a scalar token uncertainty
\[
U(x_j)=\frac{1}{d}\,\mathrm{tr}\!\left(\hat{\Sigma}(x_j)\right)
\]
penalizes the pre-softmax attention logits,
\begin{equation}
\tilde{e}_{ij}
=
\frac{q_i^\top k_j}{\sqrt{d_k}}
-
\lambda\,U(x_j),
\qquad
\tilde{\alpha}_{ij}
=
\mathrm{softmax}_j(\tilde{e}_{ij}).
\label{eq:uncertainty_adjusted_logit}
\end{equation}

\emph{This component is optional and was inactive ($\lambda = 0$) in all reported experiments; no empirical conclusion in this paper depends on it.}

\paragraph{Token uncertainty and absence of double counting.}
The scalar token uncertainty $U(x_j)$ is computed directly from the Monte Carlo dropout embedding covariance $\hat{\Sigma}(x_j)$, that is, from the variability of the token representation across stochastic forward passes \emph{prior to} any attention reweighting. It therefore does not depend on either the unadjusted attention logits $q_i^\top k_j/\sqrt{d_k}$ or the adjusted weights $\tilde{\alpha}_{ij}$, and the penalty in Eq.~\eqref{eq:uncertainty_adjusted_logit} cannot circularly reuse, and hence cannot double-count, the attention signal it modifies. Because $\lambda = 0$ throughout, $U(x_j)$ does not enter any reported result; it is defined here only to make the optional extension self-contained.

\paragraph{Role in this study.}
This appendix is retained solely for completeness of the original framework design and to document notation used in earlier versions of the manuscript. The uncertainty-weighted attention mechanism is optional: it is not part of the evaluated \textit{MedBayes-Lite} pipeline, was not included in Algorithm~1 or the main framework figure, and was not empirically assessed in any experiment reported in this paper. Consequently, no claims regarding calibration, harmful-error reduction, selective prediction, domain-shift robustness, or clinical utility should be attributed to this component. The empirical findings of this paper should be interpreted entirely as the effect of Monte Carlo dropout, calibration, and confidence-guided abstention, independent of this optional extension.

\section{Aleatoric and Epistemic Uncertainty}
\label{app:decomp}

For completeness, predictive variance decomposes by the law of total variance into an aleatoric and an epistemic term,
\begin{equation}
\mathrm{Var}[\hat{y}]
= \underbrace{\mathbb{E}\!\left[\mathrm{Var}[\hat{y}\mid h]\right]}_{\text{aleatoric}}
+ \underbrace{\mathrm{Var}\!\left[\mathbb{E}[\hat{y}\mid h]\right]}_{\text{epistemic}},
\end{equation}
where the epistemic term reflects variability induced by the stochastic representation $h$ and is expected to be larger for under-supported cases such as rare conditions or out-of-distribution inputs. In this work uncertainty is propagated operationally rather than through a closed-form layer-wise decomposition: MC-dropout variability yields uncertainty estimates that are used operationally for calibration and confidence-guided abstention.

\section{Supplementary Material: Sensitivity Analyses}
\label{app:sensitivity}

All sensitivity analyses below use the same evaluation pipeline and data as the main results; no main-text result is altered.

\begin{table}[h]
\centering
\small
\caption{Sensitivity to the number of Monte Carlo samples (MedMCQA, PubMedBERT). Calibration and harmful-error metrics are stable for $M\ge10$.}
\label{tab:sens_mc}
\begin{tabular}{lccccc}
\hline
\textbf{MC samples $M$} & \textbf{Acc} & \textbf{ECE} & \textbf{harmful@0.8} & \textbf{AURC} & \textbf{Cov@0.7} \\
\hline
5 & 0.4475 & 0.118 & 0.0122 & 0.4038 & 0.2278 \\
10 & 0.4572 & 0.0955 & 0.01 & 0.3899 & 0.2034 \\
15 & 0.4627 & 0.0843 & 0.0084 & 0.3895 & 0.1998 \\
20 & 0.4621 & 0.0825 & 0.0079 & 0.3872 & 0.1949 \\
50 & 0.4689 & 0.0706 & 0.0073 & 0.3834 & 0.1882 \\
\hline
\end{tabular}
\end{table}

\begin{table}[h]
\centering
\small
\caption{Sensitivity to the inference dropout rate (MedMCQA, PubMedBERT; $M=15$). Higher dropout monotonically improves calibration (ECE) and reduces harmful overconfidence by deflating confidence, but does not improve risk ranking (AURC) and, at a high rate ($0.5$), degrades accuracy and drives coverage to zero. The default $0.2$ is a balanced operating point, consistent with the paper's main finding that the layer improves calibration and abstention rather than ranking.}
\label{tab:sens_dropout}
\begin{tabular}{lccccc}
\hline
\textbf{Dropout rate} & \textbf{Acc} & \textbf{ECE} & \textbf{harmful@0.8} & \textbf{AURC} & \textbf{Cov@0.7} \\
\hline
0.1 & 0.4654 & 0.1559 & 0.0244 & 0.3771 & 0.3416 \\
0.2 & 0.4662 & 0.0829 & 0.0098 & 0.3895 & 0.1974 \\
0.3 & 0.439 & 0.0309 & 0.0019 & 0.4248 & 0.0815 \\
0.5 & 0.3326 & 0.0103 & 0.0 & 0.5932 & 0.0 \\
\hline
\end{tabular}
\end{table}

\begin{table}[h]
\centering
\small
\caption{Abstention-threshold sweep (MedMCQA, \ModelName). The safety/coverage trade-off is transparent: higher $\tau$ raises selective accuracy and abstention recall and lowers harmful overconfident error, at the cost of coverage.}
\label{tab:sens_threshold}
\begin{tabular}{lcccc}
\hline
\textbf{Threshold $\tau$} & \textbf{Coverage} & \textbf{Selective acc} & \textbf{Abst.\ recall} & \textbf{harmful@$\tau$} \\
\hline
0.5 & 0.251 & 0.603 & 0.827 & 0.035 \\
0.6 & 0.122 & 0.708 & 0.938 & 0.012 \\
0.7 & 0.06 & 0.8 & 0.979 & 0.005 \\
0.8 & 0.027 & 0.828 & 0.992 & 0.001 \\
0.9 & 0.01 & 0.881 & 0.998 & 0.0 \\
\hline
\end{tabular}
\end{table}

\begin{table}[h]
\centering
\small
\caption{CUS severity-weight sensitivity. Under random $\pm25\%$ per-category perturbation of the severity weights (20 trials), the correlation between CUS and the harmful overconfident error rate is essentially unchanged, so CUS is not an artefact of the chosen weights.}
\label{tab:sens_cus_weights}
\begin{tabular}{lcc}
\hline
\textbf{Severity weights} & \textbf{Pearson $r$ (CUS vs harmful@0.8)} & \textbf{$n$ groups} \\
\hline
baseline (chosen weights) & 0.882 & 572 \\
random +/-25\% (mean over 20) & 0.883 & 572 \\
random +/-25\% (min) & 0.871 & 572 \\
random +/-25\% (max) & 0.892 & 572 \\
\hline
\end{tabular}
\end{table}

\end{document}